\documentclass{article}

\PassOptionsToPackage{numbers, sort&compress}{natbib}

\usepackage[final,nonatbib]{neurips_2021}

\usepackage[utf8]{inputenc} 
\usepackage[T1]{fontenc}    
\usepackage{hyperref}       
\usepackage{url}            
\usepackage{booktabs}       
\usepackage{amsfonts}       
\usepackage{nicefrac}       
\usepackage{microtype}      
\usepackage{xcolor}         

\usepackage{amsmath,wasysym}
\usepackage[ruled,vlined,linesnumbered, resetcount]{algorithm2e}
\usepackage{graphicx}

\usepackage{subcaption}
\usepackage{multirow}
\usepackage[normalem]{ulem}
\useunder{\uline}{\ul}{}
\usepackage{textcomp}
\usepackage{footnote}
\usepackage{makecell}
\usepackage{xspace}
\usepackage{enumitem}
\usepackage{wrapfig}
\usepackage{amsthm}

\usepackage[backend=biber,maxbibnames=99]{biblatex}
\bibliography{refs}

\newcommand{\zc}[1]{\textcolor{red}{[Chao: #1]}}

\newcommand{\model}{\textsc{EpiFNP}\xspace}
\newcommand{\data}{\mathcal{D}\xspace}

\title{When in Doubt: Neural Non-Parametric Uncertainty Quantification for Epidemic Forecasting}
\author{%
	Harshavardhan Kamarthi \quad Lingkai Kong \quad Alexander Rodr\'iguez\vspace{2pt}\\
	\textbf{Chao Zhang} \quad\textbf{B. Aditya Prakash}\\
College of Computing\\
  Georgia Institute of Technology\\
	\texttt{\{harsha.pk,lkkong,arodriguezc,chaozhang,badityap\}@gatech.edu} \\
}

\begin{document}
\maketitle

\begin{abstract}

  Accurate and trustworthy epidemic forecasting is an important problem 
  for  public health planning and disease mitigation. Most existing
  epidemic forecasting models
  disregard uncertainty quantification, resulting in mis-calibrated
  predictions. Recent works in deep neural models for uncertainty-aware
  time-series forecasting also have several limitations; \textit{e.g.}, it is
  difficult to specify proper priors in Bayesian NNs, while methods like deep
  ensembling can be computationally expensive. In this paper, we propose to
  use neural functional processes to fill this gap. We model epidemic
  time-series with a probabilistic generative process and propose a functional
  neural process model called \model, which directly models the probability
  distribution of the forecast value in a non-parametric way. In \model, we
  use a dynamic stochastic correlation graph to model the correlations between
  sequences, and design different stochastic latent variables to capture
  functional uncertainty from different perspectives. Our experiments in a
  real-time flu forecasting setting show that \model significantly outperforms
  state-of-the-art models in both accuracy and calibration metrics, up to
  \emph{2.5x} in accuracy and \emph{2.4x} in calibration. Additionally, as
  \model learns the relations between the current season and similar patterns
  of historical seasons, it enables interpretable forecasts. Beyond epidemic
  forecasting, \model can be of independent interest for advancing uncertainty
  quantification in deep sequential models for predictive
  analytics. 


\end{abstract}

\section{Introduction}

\label{sec:intro}
Infectious diseases like seasonal influenza and COVID-19 are major global health issues, affecting millions of people \cite{holmdahl_wrong_2020,reich_collaborative_2019}. Forecasting  disease time-series (such as infected cases) at various temporal and spatial resolutions is a non-trivial and important task~\cite{reich_collaborative_2019}. Estimating various indicators e.g. future incidence, peak time/intensity and onset, gives policy makers valuable lead time to plan interventions and optimize supply chain decisions, as evidenced by various Centers for Disease Control (CDC) prediction initiatives for diseases like dengue, influenza and COVID-19~\cite{ray2020ensemble,johansson2019open, qian2020and}.


Statistical approaches~\cite{brooks_nonmechanistic_2018} for the forecasting problem are fairly new compared to more traditional mechanistic approaches~\cite{hethcote_mathematics_2000,shaman2012forecasting}. While valuable
for `what-if’ scenario generation, mechanistic models have several issues in real-time forecasting. For example, they
cannot easily leverage data from multiple indicators or predict composite signals. In contrast,  deep learning approaches in this context
are a novel direction and have become increasingly promising, as they can ingest numerous data signals without laborious feature
engineering~\cite{rodriguez_deepcovid_2021,ray2020ensemble,adhikari_epideep_2019,covid_fbsurvey_2020}.


However, there are several
challenges in designing such methods, primarily with the need to handle uncertainty to give more reliable forecasts~\cite{holmdahl_wrong_2020}.
Decision makers need to understand the inherent uncertainty in the forecasts so
that they can make robust decisions~\cite{ray_infectious_2017}. Providing
probabilistic forecasts and interpreting what signals cause the model uncertain
is also helpful to better communicate the situation to the public. Due to the
inherent complexity of the prediction problem, just like weather forecasting,
so-called `point' forecasts without uncertainty are increasingly seen as not
very useful for planning for such high-stake
decisions~\cite{holmdahl_wrong_2020,ray2020ensemble}.

Uncertainty quantification in purely statistical epidemic forecasting models is a little explored area. Most traditional methods optimize for accuracy of `point-estimates' only.
Some approaches that model the underlying generative distribution of the data
naturally provide a probability distribution of the outputs
\cite{brooks_flexible_2015,brooks_nonmechanistic_2018,zimmer_influenza_2020,ray_infectious_2017},
but they do not focus on producing \emph{calibrated}
distributions~\cite{guo2017calibration, pmlr-v80-kuleshov18a} as well. Another
line of research addresses this problem with the use of simple methods such as
an ensemble of models to build a sample of forecasts/uncertainty
bounds~\cite{reich_collaborative_2019, chakraborty_forecasting_2014}. Recent
attempts for deep learning forecasting models use ad-hoc methods such as
bootstrap sampling~\cite{rodriguez_deepcovid_2021};
while others disregard this aspect~\cite{wang2019defsi,rodriguez_steering_2020}. As a result these can produce wildly wrong
predictions (especially in novel/atypical scenarios) and can be even confident
in their mistakes. In time-series analysis, while a large number of deep
learning models~\cite{adhikari_epideep_2019} have been proposed, little work has
been done to quantify uncertainty in their predictions. Bayesian deep
learning~\cite{mackay1992practical, blundell2015weight,
  louizos2017multiplicative} (and approximation
methods~\cite{gal_dropout_2016,li2016preconditioned, Zhang2020Cyclical}) and
deep ensembling~\cite{lakshminarayanan2017simple} are two directions that may
mitigate this issue, but their applicability and effectiveness are still largely
limited by factors such as intractable exact model
inference~\cite{blundell2015weight, louizos2017multiplicative}, difficulty of
specifying proper parameter priors~\cite{louizos_functional_2019}, and
uncertainty underestimation~\cite{kong2020sde,kongca2020}. 
Neural Process (NP)~\cite{garnelo_neural_2018} and Functional Neural Process
(FNP)~\cite{louizos_functional_2019} are recent frameworks developed to
incorporate stochastic processes with DNNs, but only for static data.


Our work aims to close these \emph{crucial} gaps from both viewpoints. We
propose a non-parametric model for epi-forecasting by `marrying' deep sequential
models with recent development of neural stochastic processes. Our model,
called \model, leverages the expressive power of deep sequential models, while
quantifying uncertainty for epidemic forecasting directly in the functional
space. We extend the idea of learning dependencies between data points
\cite{louizos_functional_2019} to sequential data, and introduce additional
latent representations for both local and global views of input sequences to
improve model calibration. We also find that the dependencies learned by
\model enable reliable interpretation of the model's forecasts.


\begin{wrapfigure}{r}{.6\linewidth}
  \centering
  \vspace{-0.2in}
  \begin{subfigure}{.46\linewidth}
    \centering
    \includegraphics[width=\linewidth]{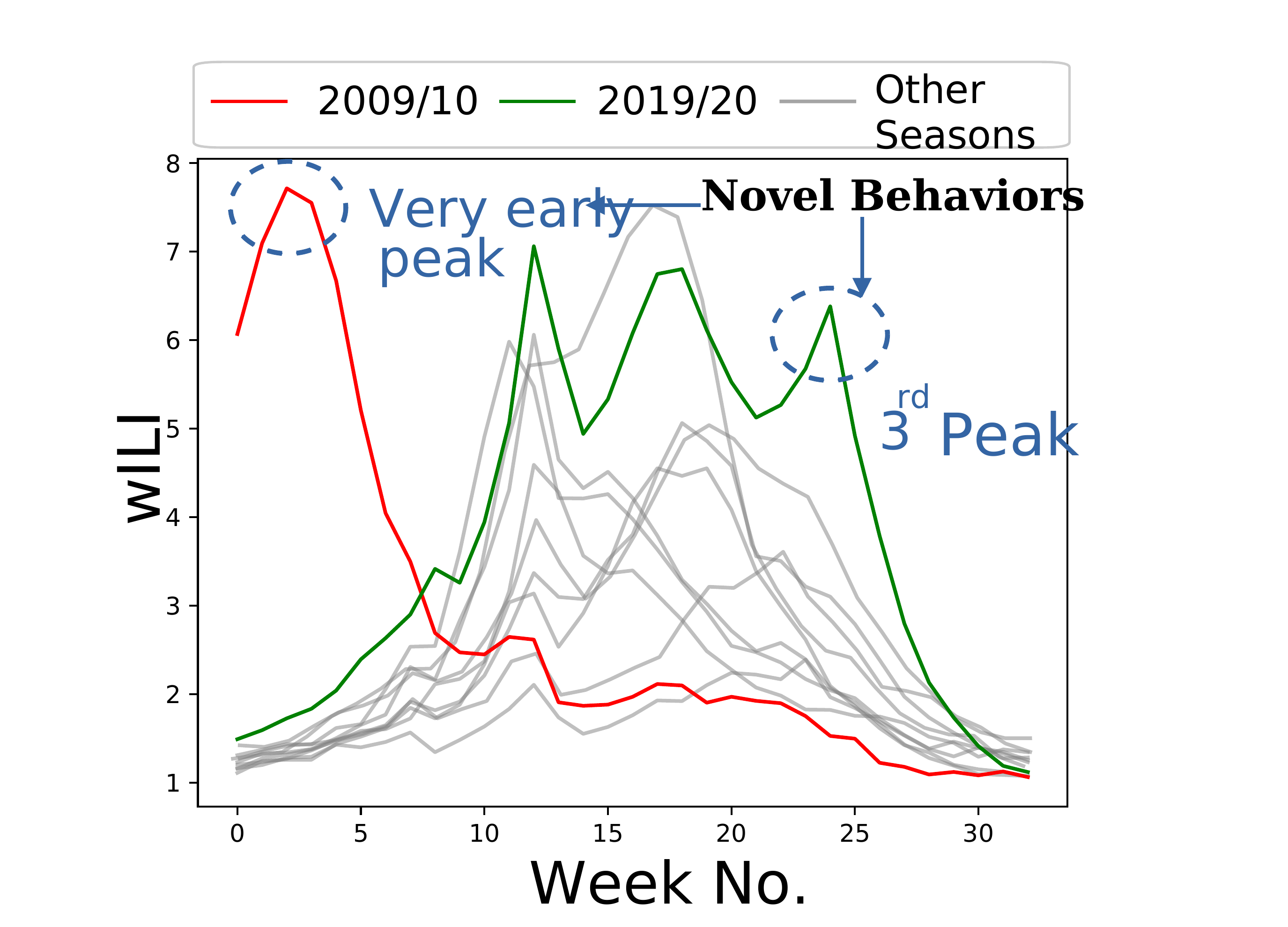}
    \caption{Historical wILI seasons sequences, 2003-20}
    \label{fig:2seasons}
  \end{subfigure}\hfill
  \begin{subfigure}{.5\linewidth}
    \centering
    \includegraphics[width=.8\linewidth]{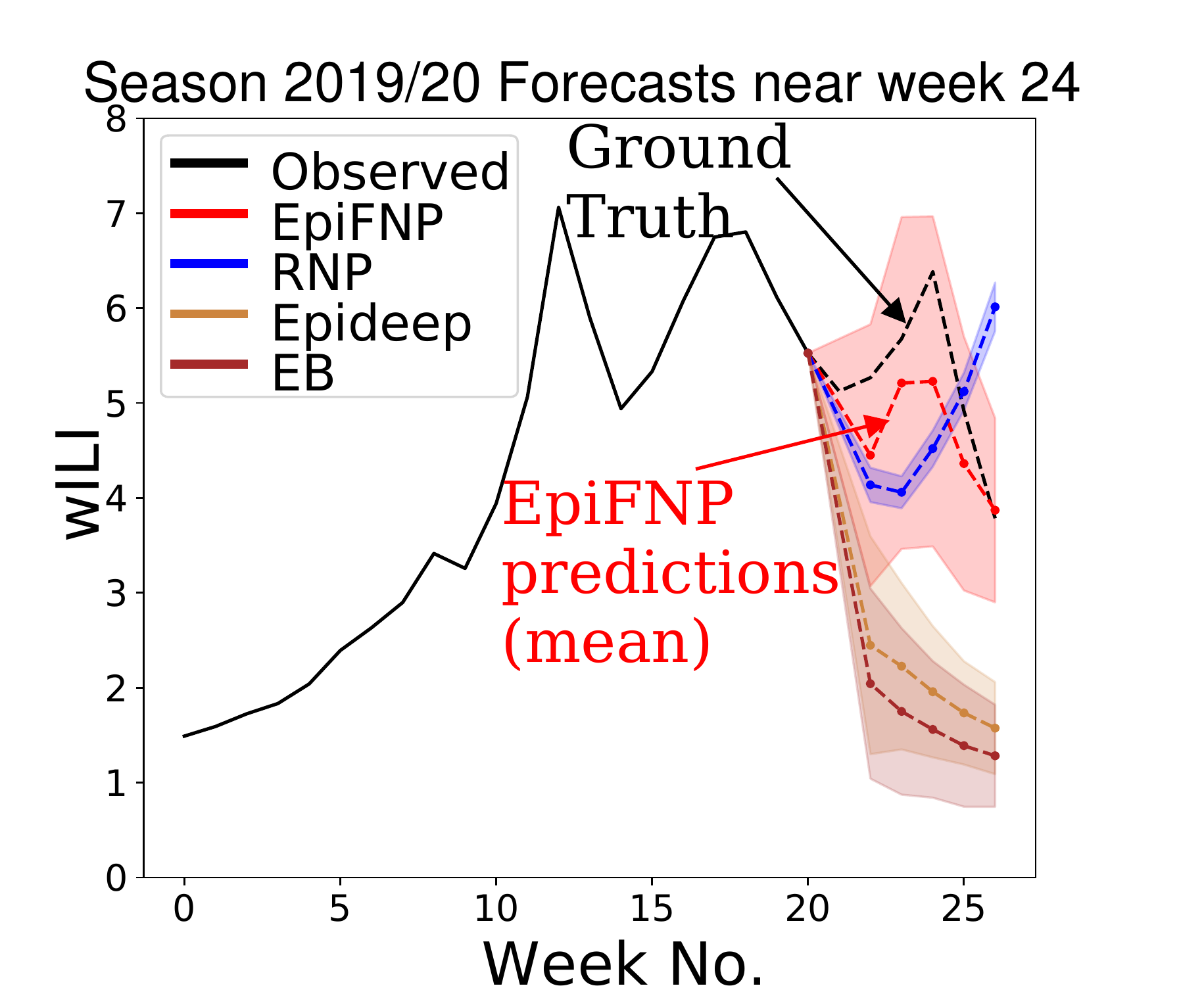}
    \caption{Probabilistic predictions of all methods}
  \end{subfigure}
  \vspace{-4pt}
  \caption{\textit{\label{fig:intro}\model(red) is the only model reacting reliably for the atypical 3rd peak of 2019/20 season and  whose 95\% confidence bounds completely encloses the ground truth.}}
\end{wrapfigure}

Figure~\ref{fig:intro} shows an example of a well-calibrated forecast due to
\model in flu forecasting. CDC is interested in forecasting weighted
Influenza-like-illness (wILI) counts, where ILI is defined
as ``fever and a cough and/or a sore throat without a known cause other than
flu. 
Figure 1 (a) shows the
historical ILI data with abnormal seasons highlighted; Figure (b) shows how our
method \model, in contrast to others, is able to react well to a particularly
novel event (in this case, introduction of a symptomatically similar COVID-19
disease), giving both \emph{accurate} and \emph{well-calibrated} forecasts.


Our main contributions are:\\
\noindent$\bullet$ \textbf{Probabilistic Deep Generative Model:} We design a  neural Gaussian processes model for epidemic forecasting,
which automatically learns stochastic correlations between query sequences and
historical data sequences for non-parametic uncertainty quantification.\\ 
\noindent$\bullet$ \textbf{Calibration and Explainability:} \model models the output forecast
distribution based on similarity between the current season and  the historical seasons in a latent space. We introduce additional latent variables to capture global information of historical seasons and local views of sequences, and show that this leads to better-calibrated forecasts. Further, the
relations learned between the current season and similar patterns  from
previous seasons enable explaining the
predictions of \model.\\ 
\noindent$\bullet$ \textbf{Empirical analysis of accurate well-calibrated
  forecasting:} We perform rigorous benchmarking on flu forecasting and  show
that \model significantly outperforms strong baselines, providing up to
\emph{2.5x} more accurate and \emph{2.4x} better calibrated forecasts. We also
use outlier seasons to show the  uncertainty in \model makes it adapt well to
unseen patterns compared with baselines.


\vspace{-5pt}
\section{Problem and Background}
\label{sec:statement}
\vspace{-5pt}

We focus on epidemic disease forecasting in this paper. Our goal is to predict the disease incidence few week into the future given the disease surveillance dataset containing incidence from the past seasons as well as for the past weeks of the current season. This is formulated as a supervised time-series forecasting problem as follows.

\noindent \textbf{Epidemic Forecasting task:} Let the incidence for season $i$ at week $t$ be $x_{i}^{(t)}$. During the current season
$N+1$ and current week $t$, we first have the snippet of time-series values
upto week $t$ denoted by $\mathbf{x}_{N+1}^{(1\dots t)} = \{x_{N+1}^{(1)},
\dots, x_{N+1}^{(t)}\}$. We are also provided with data from \emph{past}
historical seasons $1$ to $N$ denoted by $H = \{\mathbf{x}_i^{(1\dots T)}\}_{i=1}^N$
where $T$ is number of weeks per season. In \emph{real-time} forecasting,
intuitively our goal is to use all the currently available data, and predict the
next few future values (usually till 4 weeks in future).
That is to predict the value $y_{N+1}^{(t)}= x_{N+1}^{(t+k)}$, $k$ week in future where $k\in \{1,2,3,4\}$ given $\mathbf{x}_{N+1}^{(1\dots t)}$ and $H$.
Formally, our task is:
\textit{given (a) the 
dataset of historical incidence sequences $H$ 
and (b) snippet of incidence for current season $N+1$ till week $t$, $x_{N+1}^{(1...t)}$, estimate an \underline{accurate} prediction for $y_{N+1}^{(t)}$ and a \underline{well-calibrated} probability distribution $\hat{p}(y_{N+1}^{(t)}| \mathbf{x}_{N+1}^{(1...t)}, H)$.} 
There are several ways to evaluate such  forecasts~\cite{tabataba_framework_2017}, which we elaborate later in our experiments.

\vspace{-6pt}
\section{Our Methodology}
\label{sec:method}


\noindent\textbf{Overview:}
\model aims to produce calibrated forecasting probabilistic distribution. One popular choice is to use BNNs \cite{blundell2015weight,fortunato2017bayesian}  which impose probability distributions for weight parameters. However, as Deep Sequential Models (DSMs) have an enormous number of uninterpretable parameters,  it is impractical to specify proper prior distributions in the parameter space. Existing works usually adopt simple distributions \cite{blundell2015weight, ritter2018a}, e.g., independent Gaussian distribution, which could severely under-estimate the true uncertainty \cite{kong2020sde}. To solve this issue, we propose EpiFNP, which combines (1) the power of DSMs in representation learning and capturing temporal correlations; and (2) the power of Gaussian processes (GPs) in non-parametric uncertainty estimation directly in the functional space similar to \cite{louizos_functional_2019}, instead of learning probability distributions for model parameters.


 During \emph{training phase} of our supervised learning task, \model is trained to predict $x_{i}^{(t+k)}$ given $x_{i}^{(1...t)}$ as input for $i\leq N$. Therefore, we define the \underline{training set} $M$ as set of partial sequences and their forecast ground truths from historical data $H$, i.e, $M = \{(\mathbf{x}_i^{(1...t)}, y_i^{(t)}): i\leq N, t+k\leq T, y_i^{(t)}=x_i^{(t+k)}\}$. For simplicity, let $\mathbf{X}_M$ be set of the partial sequences in $M$ and $\mathbf{y}_M$ the set of ground truth labels.  Following GPs for non-parametric uncertainty quantification, \model constructs the forecasting distribution on the historical sequences. 
Since the number of possible sequences that can be extracted from $H$ is prohibitively large, we narrow down the set of candidates into a set of sequences that comprehensively represents $H$, called the \underline{reference set} $R$. 
We choose the set of full sequences of $T$ incidence values for each season as reference set, i.e,  $R = \{\mathbf{x}_{i}^{(1...T)}\}_{i=1}^{N_R}$. 
We refer elements of $M$ as $\{\mathbf{x}_i^{M} ,y_i^{M}\}_{i=1}^{N_M}$ and $R$ as $\{\mathbf{x}_i^{M}\}_{i=1}^{N_R}$
when we don't need to specify the week and season. 
Also let $\mathbf{X}_\data= \{\mathbf{x}_i^{M}\}_{i=1}^{N_{M}} \cup \{\mathbf{x}_i^{R}\}_{i=1}^{N_{R}} $, the union of reference and training sequences.


The generative process of \model includes three key steps (also see Figure~\ref{fig:arch} and Eq.~\ref{likelihood}):

\begin{figure}[t]
    \centering
    \includegraphics[width=.9\linewidth]{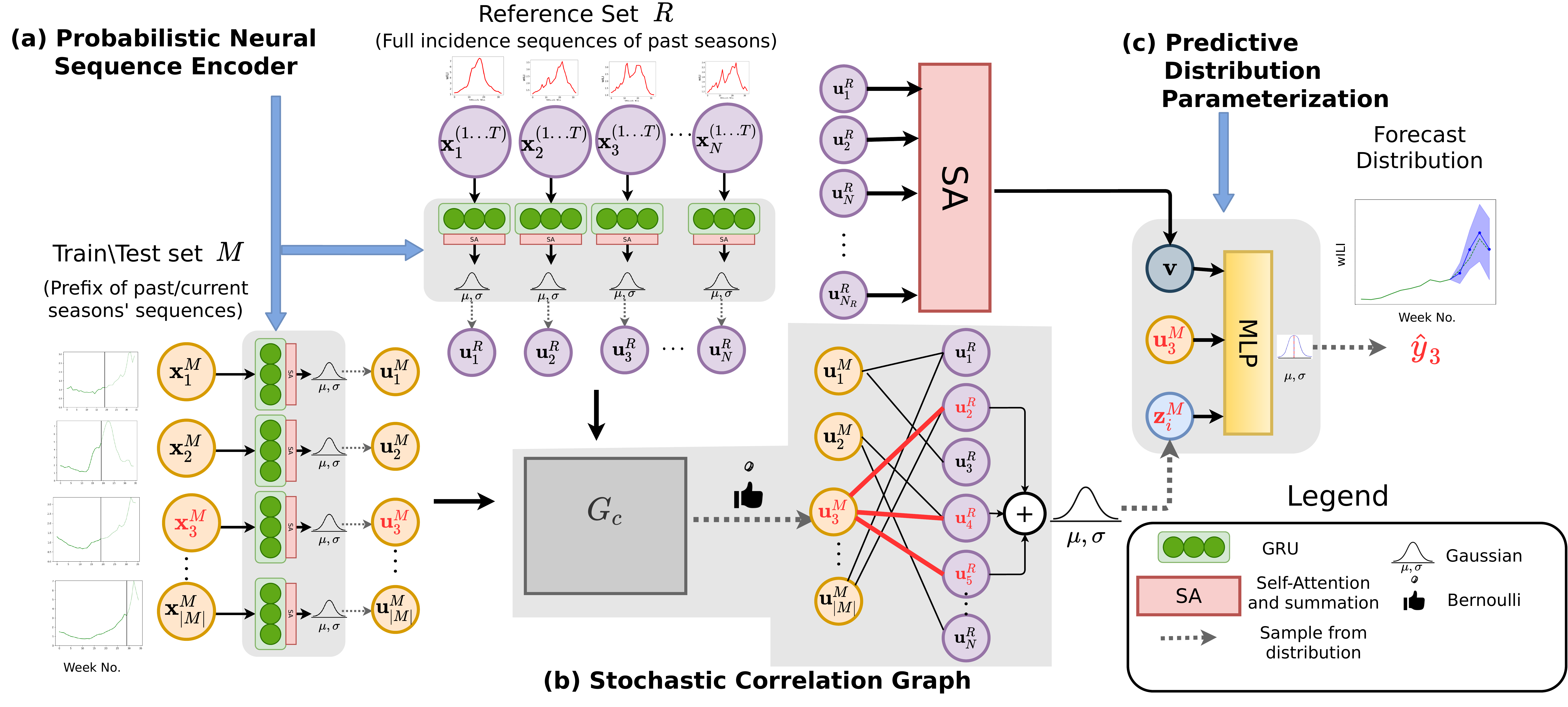}
    \caption{\textit{Pipeline of proposed \model model. (i) Three main components (a), (b) and (c) correspond to the terms in Equation~\ref{likelihood}. (ii) Variables highlighted in {\color{red} Red} correspond to steps specific to inference of sequence $\mathbf{x}_{3}^M$. \label{fig:arch}}}
\vspace{-10pt}
\end{figure}

\begin{enumerate}[label=(\alph*), wide, labelwidth=!, labelindent=0pt,partopsep=0pt,topsep=0pt,parsep=0pt]
    \item \textbf{Probabilistic neural sequence encoding} (Section 4.2). 
The first step of the generative process is to use a DSM to encode the sequence $\mathbf{x}_i \in \mathbf{X}_\data$
into a
\textit{variational} latent embedding $\mathbf{u}_i \in \mathbf{U}_\data$.
The representation power of DSM helps us to
model complex temporal patterns within sequences, while the probabilistic encoding framework enables us
to capture the uncertainty in sequence embedding. 

\item \textbf{Stochastic correlation graph construction} (Section 4.3). The second step  is to capture the correlations between reference ($\mathbf{U}_R$) and training ($\mathbf{U}_M$) data points in the \emph{latent embedding space}  (i.e. seasonal similarity between epidemic curves).
We use a stochastic data correlation graph $\mathbf{G}$, which plays a similar role
to the covariance matrix in classic GPs. It encodes the dependencies between reference  and
training sequences, enabling non-parametric uncertainty estimation.

\item \textbf{Final predictive distribution parameterization} (Section 4.4).
Finally, we parameterize the predictive distribution with three stochastic latent variables:
(1) The global stochastic latent variable $\mathbf{v}$, which is shared by all the sequences. This variable captures the overall information of the underlying function based on all the reference points. (2) The local stochastic latent variables $\mathbf{Z}_{M}=\{\mathbf{z}_i^M \}_{i=1}^{N_M}$. This term captures the data correlation uncertainty based on the stochastic data correlation graph $\mathbf{G}$. (3) The stochastic sequence embeddings $\mathbf{U}_{M}=\{\mathbf{u}^M_i
\}_{i=1}^{N_M}$. This term captures the embedding uncertainty and provide additional information beyond the reference set.


\end{enumerate}

Hence, putting it all together from the generative process, we factorize the predictive distribution of the training sequences into three corresponding parts ($\theta$ is the union of the parameters in \model): 
{\small
\begin{equation}
   \begin{split}
    & p(\mathbf{y}_M|\mathbf{X}_M, R)  = \sum_{\mathbf{G}} \int  \underbrace{p_{\theta}(\mathbf{U}_{\mathcal{D}}|\mathbf{X}_{\mathcal{D}})}_{\text{(a)}} \underbrace{p(\mathbf{G}|\mathbf{U}_{\mathcal{D}})}_{\text{(b)}} \\
    &\underbrace{p_{\theta}(\mathbf{Z}_{M}, |\mathbf{G},\mathbf{U}_{R})p_{\theta}( \mathbf{v}|\mathbf{U}_{R})p_{\theta}(\mathbf{y}_M|\mathbf{U}_{M}, \mathbf{Z}_M, \mathbf{v})}_{\text{(c)}}
    d\mathbf{U}_{\mathcal{D}} d\mathbf{Z}_M d\mathbf{v}.
    \end{split}
    \label{likelihood}
\end{equation}
}
\noindent
Compared to existing recurrent neural process (RNP) \cite{qin_recurrent_2019} for sequential data (and its related predecessors~\cite{garnelo_neural_2018,kim2019attentive}), our \model process has stronger representation power
and is more interpretable. Specifically,  RNP uses a single global stochastic latent variable to capture the
functional uncertainty, which is not flexible enough to represent a complicated underlying
distribution. In contrast, \model constructs three stochastic latent variables to capture the
uncertainty from different perspectives and can interpret the prediction based on the correlated
reference sequences. 
\vspace{-5pt}
\subsection{Probabilistic Neural Sequence Encoder}
\label{sec:seqs}
\vspace{-5pt}
The probabilistic neural sequence encoder $p_{\theta}( \mathbf{U}_D |\mathbf{X}_D )$ aims to model
the complex temporal correlations of the sequence for accurate predictions of $y$, while capturing the
uncertainty in the sequence embedding process. To this end, we design the \emph{sequence encoder} as a
RNN and stack \emph{a self-attention layer} to capture long-term correlations. Moreover, following Variational
auto-encoder (VAE) \cite{kingma2013auto}, we model the latent embedding $\mathbf{u}_i$ as a Gaussian random variable to
capture embedding uncertainty.


We encode all the sequences, including reference sequences and training sequences, independently. Taking one sequence $\mathbf{x}_i$ as an example,
we first feed $\mathbf{x}_i$
into a Gated Recurrent Unit (GRU) \cite{cho2014learning}:
\begin{equation}
     \{\mathbf{h}_{i}^{(1)}\dots, \mathbf{h}_{i}^{(t)}\} = \text{GRU}(\{x_{i}^{(1)}\dots, x_{i}^{(t)}\}).
\end{equation}
where $\mathbf{h}_i^{(t)}$ denotes the hidden state at time step $t$.
To obtain the embedding of $\mathbf{x}_i$, the simplest way is to directly use the last step hidden
state, $\mathbf{h}^{(t)}$. However, using the last step embedding is inadequate for
epidemic forecasting as the estimates for ILI surveillance data are often delayed and revised
multiple times before they stabilize \cite{adhikari_epideep_2019}. Over-reliance over the last step hidden state would harm
the predictive ability of the model. Therefore, we choose to use a self-attention layer \cite{vaswani2017attention} to
aggregate the information of
the hidden states across all the time steps:
\begin{equation}
 \{{\alpha}_{i}^{(1)}\dots, {\alpha}_{i}^{(t)}\}  = \text{Self-Atten}(\{{\mathbf{h}}_{i}^{(1)}\dots, {\mathbf{h}}_{i}^{(t)}\}),
 \quad\quad
 \bar{\mathbf{h}}_{i} = \sum_{t'=1}^t \alpha_{i}^{(t')}\mathbf{h}_i^{(t')},
\end{equation}
where $\bar{\mathbf{h}}_{i}$ is the summarized hidden state vector. Compared with the vanilla attention  mechanism  \cite{bahdanau2015neural}, self-attention is better at capturing long-term temporal correlations \cite{vaswani2017attention}.
Though $\bar{\mathbf{h}}_{i}$ has encoded the temporal correlations, it is deterministic and cannot represent embedding uncertainty. Inspired by VAE, we parameterize each dimension of the latent embedding $\mathbf{u}_i$ as a Gaussian random variable:
\begin{equation}
\label{eqn:stoseq}
    p_{\theta}([\mathbf{u}_{i}]_k|\mathbf{x}_i) = \mathcal{N}([g_1(\bar{\mathbf{h}}_{i})]_k, \exp([g_2(\bar{\mathbf{h}}_{i})]_k)),
\end{equation}
where $g_1$ and $g_2$ are  two multi-layer perceptrons (MLPs), $[\cdot]_k$ is the $k$-th dimension of the variable.
\vspace{-5pt}
\subsection{Stochastic Data Correlation Graph}
\label{sec:relation}
\vspace{-5pt}
The stochastic graph $\mathbf{G}$ is used to model the correlations among sequences, which is central to
the non-parametric uncertainty estimation ability of \model. It is realized by
 constructing a bipartite graph from the reference set $R$ to the training set $M$ \emph{based on the
 similarity between their sequence embeddings}.
With this graph, we aim to model the dynamic similarity among epidemic curves as in \cite{adhikari_epideep_2019} but in a stochastic manner, which allows us to further quantify the uncertainty coming from our latent representations of the sequences. Note that the similarity with reference sequence embeddings dynamically changes across the current season since different periods of the season may be similar to different sets of reference sequences (as we illustrate in Section~\ref{subsec:explainable}).

We first construct a complete weighted bipartite graph $\mathbf{G}_c$ from $R$ to $M$, where the nodes are the sequences. \emph{The weight of each edge is calculated 
as  similarity between two sequences in the embedding space using the radial basis function kernel} $\kappa(\mathbf{u}^R_i, \mathbf{u}^M_j)=\exp(-\gamma ||\mathbf{u}^R_i -\mathbf{u}^M_j||^2)$. Modeling such a similarity in the embedding space is more accurate than in the input space by leveraging the representation power of the neural sequence encoder.

Though we can directly use $\mathbf{G}_c$ to encode the data correlations, such a dense complete graph
\begin{wrapfigure}{r}{.5\linewidth}
\vspace{-0.2in}
    \centering
    \includegraphics[width=0.9\linewidth]{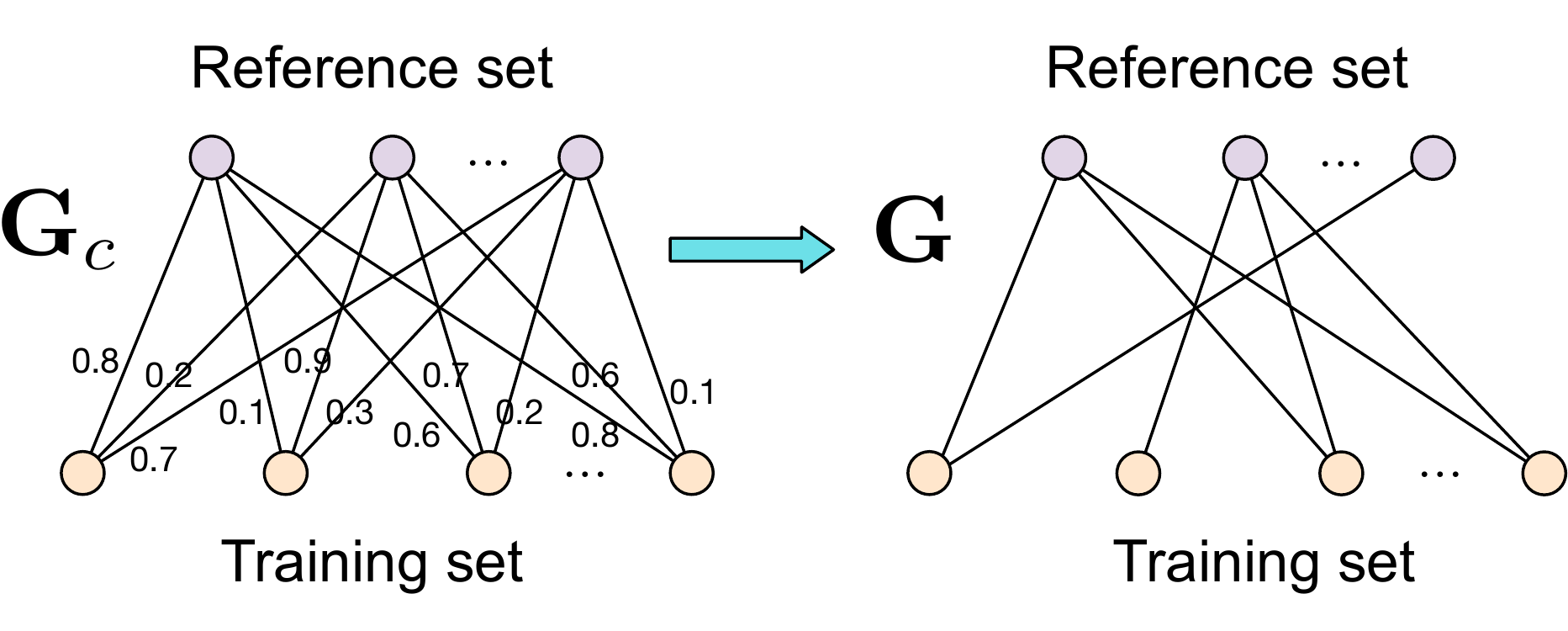}
    \vspace{-0.2cm}
    \caption{\textit{We sample the (sparse) binary graph $\mathbf{G}$ from the complete weighted (dense) graph $\mathbf{G}_c$.} \label{fig:graph}}
    \vspace{-0.5cm}
\end{wrapfigure}
 requires heavy computations and does not scale to a large
dataset.  Therefore, we choose to further sample from this complete graph to obtain a stochastic binary bipartite graph $\mathbf{G}$ as shown in Figure~\ref{fig:graph}.  This graph can be represented as a random binary adjacency matrix, where $\mathbf{G}_{i,j}=1$ means the reference sequence $\mathbf{x}_i^R$ is a parent of the training sequence $\mathbf{x}_j^M$.
We then parameterize this binary adjacency matrix using
Bernoulli distributions:
\begin{equation}
    p(\mathbf{G}|\mathbf{U}_{\mathcal{D}}) =\prod_{i\in R}\prod_{j \in M} \text{Bernoulli}(\mathbf{G}_{i,j}|\kappa(\mathbf{u}_i^R, \mathbf{u}_j^M)).
\end{equation}
 Intuitively, the edges in $\mathbf{G}_c$ with higher weights are more likely to be kept after
 sampling. This sampling process leads to sparse correlations for each sampled graph,  which can
 speed up training due to sparsity.
\vspace{-5pt}
\subsection{Parameterizing Predictive Distribution}
\vspace{-5pt}
Here we introduce how to parameterize the final prediction based on the three latent variables mentioned in Section 4.1, which \emph{capture the functional uncertainty from different perspectives}.

\noindent
\textbf{Local latent variable $\mathbf{z}_{i}^M$}: It summarizes the information of the correlated reference points for each training point and captures the \emph{uncertainty of data correlations}. We generate $\mathbf{z}_{i}^M$ based on the structure of the data correlation graph, and  each dimension $k$ follows a Gaussian distribution:
\begin{equation}
    \mathbf{z}_{i,k}^M \sim \mathcal{N}(C_i\sum_{j:\mathbf{G}_{j,i}=1} h_1(\mathbf{u}_j^R)_k, \exp(C_i\sum_{j:\mathbf{G}_{j,i}=1} h_2(\mathbf{u}_j^R)_k)),
    \label{eq:local}
\end{equation}
where $h_1$ \& $h_2$ are two MLPs and $C_i=\sum_{j}\mathbf{G}_{i,j}$ is for normalization.
As we can see from Equation~\ref{eq:local}, if the sequence has lower probability to be connected
with the reference sequences, $\mathbf{z}_{i}^M$ becomes a standard Gaussian distribution which is
an uninformative prior. This property imposes a similar inductive bias as in the GPs with RBF kernel.

\noindent
\textbf{Global latent variable $\mathbf{v}$}. It encodes the
information in \emph{all the reference points}, computed as:
\begin{equation}
     \beta_1, \dots, \beta_{N_R} = \text{Self-Atten}(\mathbf{u}^R_1, \dots, \mathbf{u}^R_{N_R}),
     \quad\quad\quad
      \mathbf{v} = \sum_{i=1}^{N_R} \beta_i\mathbf{u}_i^R.
\end{equation}
In contrast with the local variable $\mathbf{z}_i^M$, the global latent variable $\mathbf{v}_i$ summarizes the overall information of the underlying function. It
is shared by all the training sequences which allows us to capture the \emph{functional uncertainty from a global level}. 


\noindent
\textbf{Sequence embedding $\mathbf{u}_i^M$}: The above two latent variables are both constructed
from the embeddings of the reference sequences, which may lose \emph{novel information present in the training sequences}.
Therefore, we add a direct path from the latent embedding $\mathbf{u}^M_i$ of the training
sequence to the final prediction to enable the neural network to extrapolate beyond
the distribution of the reference sequences. This is useful in novel/unprecedented patterns where the input sequence can not rely only on reference sequences from historical data for prediction.

We concatenate the three  variables together into a single vector $\mathbf{e}_i$ and obtain the final predictive distribution (where $d_1$ and $d_2$ are MLPs):
\begin{equation}
    \mathbf{e}_i = \text{concat}(\mathbf{z}_i, \mathbf{v}_i, \mathbf{u}_i),
    \quad\quad\quad
     p(y_i| \mathbf{z}_i^M, \mathbf{v}, \mathbf{u}_i^M) = \mathcal{N}(d_1(\mathbf{e}_i), \exp(d_2(\mathbf{e}_i))).
\end{equation}
\subsection{Learning the distribution}
\vspace{-5pt}
We now introduce how to learn the model parameters  efficiently during training and forecast for a new unseen
sequence at test time.
Directly maximizing the data likelihood is intractable due to the summation and integral in Equation~\ref{likelihood}. Therefore, we choose to use the \emph{amortized variational inference} and approximate the true posterior $p(\mathbf{U}_{\mathcal{D}},\mathbf{G},\mathbf{Z}_M,\mathbf{v}|R, M)$ with $q_\phi(\mathbf{U}_{\mathcal{D}},\mathbf{G},\mathbf{Z}_{M},\mathbf{v}|R, M)$, similar to \cite{louizos_functional_2019}, as
\vspace{-5pt}
\begin{equation}
    q_\phi(\mathbf{U}_{\mathcal{D}},\mathbf{G},\mathbf{Z}_M,\mathbf{v}|R, M) = p_{\theta}(\mathbf{U}_{\mathcal{D}}|\mathbf{X}_{\mathcal{D}})p(\mathbf{G}|\mathbf{U}_{\mathcal{D}})p(\mathbf{v}|\mathbf{U}_{R}) q_\phi(\mathbf{Z}_M|M).
\label{eqn:qdist}
\end{equation}
We design  $q_{\phi}$ as a single layer of neural network parameterized by $\phi$, which outputs
mean and variance of the Gaussian distribution $q_{\phi}(\mathbf{Z}_{M}|\mathbf{X}_{M})$.

We then use a gradient-based method, such as Adam~\cite{kingma2013auto}, to maximize the evidence lower bound (ELBO) of
the log likelihood. After canceling redundant terms, the ELBO can be written as:
\begin{equation}
\begin{split}
    \mathcal{L} = -\mathrm{E}_{\mathbf{Z}_M,\mathbf{G},\mathbf{U}_{\mathcal{D}}, \mathbf{v} \sim q_{\phi}(\mathbf{Z}_M|\mathbf{X}_{M})p_{\theta}(\mathbf{G},\mathbf{U}_{\mathcal{D}},\mathbf{v}|\mathcal{D})} & [ \log P(\mathbf{y}_M|\mathbf{Z}_M,\mathbf{U}_{M}, \mathbf{v}) \\+ \log P(\mathbf{Z}_M|\mathbf{G},\mathbf{U}_{R}) - q_{\phi}(\mathbf{Z}_M|\mathbf{X}_{M})  ].
\end{split}
\end{equation}
We use the reparameterization trick to make the sampling procedure from the Gaussian distribution
differentiable. Moreover, as sampling from the Bernoulli distribution in Equation 7 leads to
discrete correlated data points, we make use of the Gumbel softmax trick \cite{jang2016categorical} to
make the model differentiable.

At test time, with the optimal parameter $\theta_{\rm opt}$, we base the predictive distribution of a new unseen partial sequence  $\mathbf{x}^{*}$  on the reference set as:
\begin{equation}
\begin{split}
    p(y^{*}|R,\mathbf{x}^{*})=&p_{\theta_{\rm opt}}(\mathbf{U}_R, \mathbf{u}^{*}|\mathbf{X}_M, \mathbf{x}^{*})p(\mathbf{a}^{*} |\mathbf{U}_R, \mathbf{u}^{*}) \\
    & p_{\theta_{\rm opt}}(\mathbf{z}^{*}|\mathbf{a}^{*}, \mathbf{U}_R, \mathbf{u}^{*})p_{\theta_{\rm opt}}(y^{*}|\mathbf{u}^{*}, \mathbf{z}^{*}, \mathbf{v})d\mathbf{U}_Rd\mathbf{z}^{*}d\mathbf{v},
\end{split}
\end{equation}
where $\mathbf{a}^{*}$ is the binary vector that denotes
which reference sequences are the parents of the new sequence. $\mathbf{u}^{*}$ and
$\mathbf{z}^{*}$ are latent embedding and local latent variable for the new sequence, respectively.
\vspace{-12pt}
\section{Experiments}
\label{sec:experiments}
\vspace{-5pt}
All experiments were done on an Intel i5 4.8 GHz CPU with Nvidia GTX 1650 GPU. The model typically takes around 20 minutes to train. The code is implemented using Pytorch and will be released for research purposes. Supplementary contains additional details and results (e.g. hyperparameters, results on additional metrics (MAPE), additional case and ablation studies). 


\noindent  \textbf{Dataset:} In our experiments, we focus on flu forecasting.
The CDC uses the ILINet surveillance system  to gather flu information from public health labs and clinical institutions across the US. 
It releases weekly estimates of \emph{weighted influenza-like illness} (wILI)\footnote{\url{https://www.cdc.gov/flu/weekly/flusight/index.html}}: out-patients with flu-like symptoms aggregated for US national and 10 different  regions (called HHS regions). Each flu season begins at week 21 and ends on week 20 of the next year e.g. Season 2003/04 begins on week 21 of 2003 and ends on week 20 of 2004. Following the guidelines of CDC flu challenge \cite{adhikari_epideep_2019,reich_collaborative_2019}, we predict from week 40 till the end of season next year. We evaluate our approach using wILI data of 17 seasons from 2003/04 to 2019/20 
.

\noindent \textbf{Goals:} Our experiments were designed evaluate the following.
    \textbf{Q1:}  Accuracy and calibration of \model's  forecasts.
    \textbf{Q2:} Importance of different components of \model.
     \textbf{Q3:} Utility of uncertainty estimates for other related tasks?.
     \textbf{Q4:} Adaptability of \model to novel behaviors during real-time forecasting.
    \textbf{Q5:} Explainability of predictions. 

\label{sec:metrics}

\noindent \textbf{Evaluation metrics:}
Let $x_{N+1}^{1...t}$ be a given partial wILI test sequence with observed ground truth $y_{N+1}^{(t)}$ i.e., for a $k$-week-ahead task $y_{N+1}^{(t)}$ is just $x_{N+1}^{(t+k)}$.
For a model/method $M$ let $\hat{p}_{N+1,M}^{(t)}(Y)$ be the output distribution of the forecast with mean $\hat{y}_{N+1,M}^{(1...t)}$.  To measure the predictive accuracy, we use \textbf{Root Mean Sq. Error} (RMSE), \textbf{Mean Abs. Per. Error} (MAPE) and \textbf{Log Score} (LS) which are commonly used in CDC challenges \cite{adhikari_epideep_2019, reich_collaborative_2019}).  To evaluate the calibration of the predictive distribution we introduce a new metric called \textbf{Calibration Score} (CS).  For a model $M$ we define a function $k_M: [0,1] \rightarrow [0,1]$ as follows. For each value of confidence $c \in[0,1]$, let $k_M(c)$ denote the fraction of observed ground truth that lies inside the $c$ confidence interval of predicted output distributions of $M$.
 For a perfectly calibrated model $M^*$ we would expect $k_{M^*}(c)=c$. CS measures the deviation of $k_{M}$ from $k_{M^*}$. Formally, we define CS as:
 {\small
\begin{equation}
    CS(M)=\int_{0}^1 |k_M(c)-c| dc \approx 0.01 \sum_{c\in\{0, 0.01, \dots, 1\}} |k_M(c)-c|.
\end{equation}
}
For all metrics, lower is better. We also define the \textbf{Calibration Plot} (CP) as the profile of $k_M(c)$ vs $c$ for all $c\in[0,1]$.

\noindent \textbf{Baselines:} \label{sssec:baselines}
We compare \model with standard and state-of-art models used for flu forecasting before, as well as methods typically used for learning  calibrated uncertainty quantification.\\
\noindent \textit{\underline{Flu forecasting related:}} $\bullet$~\textbf{SARIMA:} Seasonal Autoregressive Integrated Moving-Average is a auto-regressive time series model used as baseline for forecasting tasks \cite{adhikari_epideep_2019, zimmer_influenza_2020}. \noindent$\bullet$~\textbf{Gated Recurrent Unit} (GRU): A popular deep learning sequence encoder, used before as a baseline for this problem~\cite{adhikari_epideep_2019}.
\noindent$\bullet$~\textbf{Empirical Bayes} (EB): Utilizes a bayes framework  and has won few epidemic forecasting competitions in past~\cite{brooks_flexible_2015}.
\noindent$\bullet$~\textbf{Delta Density} (DD): A probabilistic modelling approach that learns distribution of change in successive wILI values given changes from past weeks~\cite{brooks_nonmechanistic_2018}.
\noindent$\bullet$~\textbf{Epideep} (ED) \cite{adhikari_epideep_2019}:  Recent state-of-the-art NN flu prediction model based on learning similarity between seasons. 
\noindent$\bullet$~\textbf{Gaussian Process} (GP)  \cite{zimmer_influenza_2020}: Recently proposed statistical flu prediction model using  GPs. Note that ED, SARIMA and GRU can only output point estimates 
and we use the ensemble approach to obtain their uncertainty estimates following \cite{reich_collaborative_2019, chakraborty_forecasting_2014}.\\
\noindent \textit{\underline{General ML Uncertainty related:}} $\bullet$ \textbf{Monte Carlo Dropout}  (MCDP) \cite{gal_dropout_2016}:  MCDP applies dropout at testing time for multiple times to measure the uncertainty. We use MCDP on a GRU as a baseline.
 \noindent$\bullet$ \textbf{Bayesian neural network }(BNN) \cite{blundell2015weight}:
    BNN imposes and learns from probability distributions over model parameters. 
    We used LSTM as the architecture for BNN
    \noindent$\bullet$
    \textbf{Recurrent Neural Process} (RNP) \cite{qin_recurrent_2019}: This method builds on Neural Process framework to learn from sequential data.

\emph{Note:} We need to train \model only once at start of a season using data from all past seasons unlike some baselines (ED, EB, GP, SARIMA, DD) which require retraining each week.

\subsection{Q1 \& Q2: Forecast Accuracy, Calibration and Model Ablation}
\label{sec:mainres}

\begin{table}[h]
\centering
\caption{Average US National Performance: $k$ week ahead forecasting for seasons 2014/15-2019/20.}
\scalebox{0.8}{\small 
   \begin{tabular}{|l|r|r|r|r|r|r|r|r|r|r|r|r|}
\hline
                               & \multicolumn{3}{c|}{{\ul \textbf{RMSE}}}      & \multicolumn{3}{c|}{{\ul \textbf{MAPE}}}         & \multicolumn{3}{c|}{{\ul \textbf{LS}}}        & \multicolumn{3}{c|}{{\ul \textbf{CS}}}           \\ \hline
{\ul \textit{Model}}           & k=2           & k=3           & k=4           & k=2            & k=3            & k=4            & k=2           & k=3           & k=4           & k=2            & k=3            & k=4            \\ \hline
\textbf{ED}                    & 0.73          & 1.13          & 1.81          & 0.14           & 0.23           & 0.33           & 4.26          & 6.37          & 8.75          & 0.24           & 0.15           & 0.42           \\ \hline
\textbf{GRU}                   & 1.72          & 1.87          & 2.12          & 0.28           & 0.31           & 0.356          & 7.98          & 8.21          & 8.95          & 0.16           & 0.2            & 0.22           \\ \hline
\textbf{MCDP}                  & 2.24          & 2.41          & 2.61          & 0.46           & 0.51           & 0.6            & 9.62          & 10            & 10            & 0.24           & 0.32           & 0.34           \\ \hline
\textbf{GP}                    & 1.28          & 1.36          & 1.45          & 0.21           & 0.22           & 0.26           & 2.02          & 2.12          & 2.27          & 0.24           & 0.25           & 0.28           \\ \hline
\textbf{BNN}                   & 1.89          & 2.05          & 2.43          & 0.34           & 0.46           & 0.51           & 6.92           & 7.56          & 8.03        & 0.18           & 0.22           & 0.25           \\ \hline
\textbf{SARIMA}                & 1.43          & 1.81          & 2.12          & 0.28           & 0.35           & 0.42           & 3.11          & 3.4           & 3.81          & 0.43           & 0.38           & 0.34           \\ \hline
\textbf{RNP}                   & 0.61          & 0.98          & 1.18          & 0.13           & 0.22           & 0.29           & 3.34          & 3.61          & 3.89          & 0.43          & 0.46          & 0.45          \\ \hline
\textbf{EB}                    & 1.21          & 1.23          & 1.25          & 0.57           & 0.58           & 0.58           & 6.92          & 7             & 7.12          &   0.07             &    0.082            &    0.085            \\ \hline
\textbf{DD}                    & 0.6           & 0.79          & 0.94          & 0.35           & 0.41           & 0.45           & 3.56          & 3.87          & 4.02          &   0.12             &   0.12             &     0.13           \\ \hline
\textbf{\model} & \textbf{0.48} & \textbf{0.79} & \textbf{0.78} & \textbf{0.089} & \textbf{0.128} & \textbf{0.123} & \textbf{0.56} & \textbf{0.84} & \textbf{0.89} & \textbf{0.068} & \textbf{0.081} & \textbf{0.035} \\ \hline
\end{tabular}
}
    \label{tab:kddpred}
\end{table}

\textbf{Prediction Accuracy:} We first compare the accuracy of \model against all baselines for real-time forecasting in Table \ref{tab:kddpred}.
  \model \textit{significantly} outperforms all other baselines for RMSE, MAPE, LS (which measure forecast accuracy). We notice around 13\% and 42\% improvement over the second best baseline in RMSE and MAPE respectively. Impressively LS of \model is \textit{2.5 to 3.5} times less than closest baseline\footnote{Our results are statistically significant using the Wilcox signed ranked test ($\alpha=0.05$) with 5 runs.}. This is because the intervals $y_i^{(t)}\pm 0.5$ of ground truth consistently fall inside high probability regions of our forecast distribution due to better accuracy (of mean) in general. Even during weeks of uncertainty (like around the peaks) most baselines badly calibrated forecasts don't sufficiently cover the interval, \model's distribution are wide enough to capture this interval thanks to its superior representation power. 
We also observed similar results for the 10 HHS regions as well where \model outperforms the baselines where we show 16\% and 7\% improvements in RMSE ans LS respectively showing \model's proficiency over large variety different regions and seasons.

\begin{wrapfigure}[13]{l}{.6\linewidth}
\vspace{-18pt}
    \centering
    \begin{minipage}{.5\linewidth}
    \includegraphics[width=.9\linewidth]{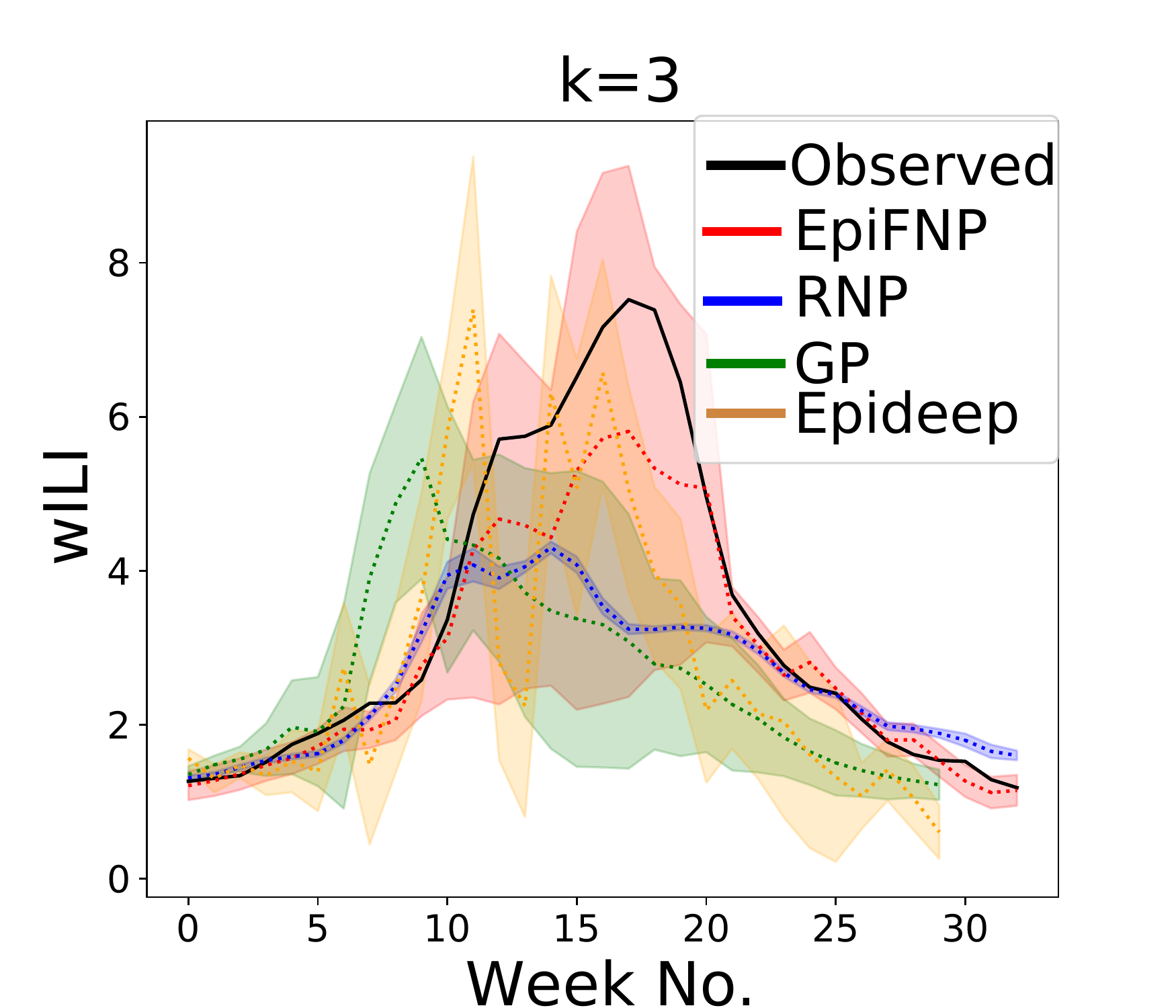}
    \caption{\textit{Forecasts and 95\% confidence bounds on  2017/18 season.}}
    \label{fig:subassembly}
    \end{minipage}\hfill
    \begin{minipage}{.46\linewidth}
      \includegraphics[width=\linewidth]{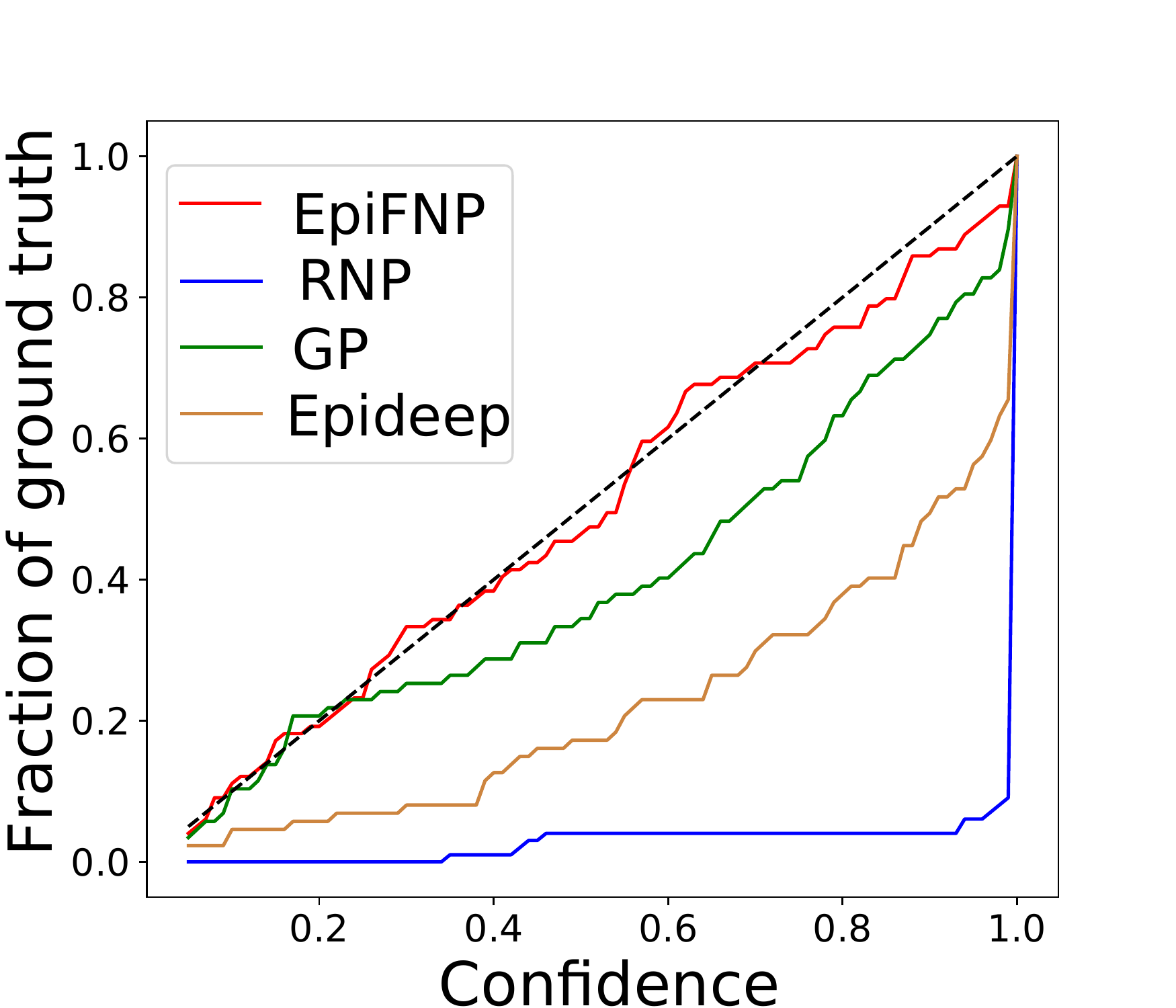}
      \caption{\textit{CPs for \model and next 3 accurate baselines, k=4}}
    \label{fig:2kddconfpl}
    \end{minipage}\vspace{-5pt}
\end{wrapfigure}
\noindent \textbf{Calibration Quality:} We measure how well-calibrated \model's uncertainty bounds (Figure \ref{fig:subassembly}) are via CS. \model was again the clear winner both for national forecasts (Table~\ref{tab:kddpred}) and regional forecasts. Calibration Plots (CPs) (Figure~\ref{fig:2kddconfpl}) show \model 
is much closer to the diagonal line (ideal calibration) compared to even the most competitive baselines. 
We also observed that applying post-hoc calibration methods \cite{kuleshov2018accurate,song2019distribution} doesn't effect the significance of \model's calibration performance (Appendix Table \ref{tab:posthoc}).
\emph{\model is clearly significantly superior to all other baselines in predicting both a  better calibrated and more accurate forecast distribution.}

\textbf{Ablation studies:}
We found all three of our \model components important for performance, with the data correlation graph the most relevant in determining uncertainty bounds. Refer to supplementary for complete results and further discussion.

\subsection{Q3: Effective uncertainty estimates: Autoregressive inference}
\label{sec:ar}
\noindent \textbf{Motivation:} We further show the usefulness and quality of our uncertainty estimates by leveraging the so-called 'auto-regressive' inference (ARI) task. It is common to perform such forecasting in real-time epidemiological settings, especially as accuracy and training data typically drops with increasing $k$ week-ahead in future~\cite{rodriguez_deepcovid_2021}. In this task, the model uses its own output for $k=1$ forecast as input (multiple samples) to predict $k=2$ forecasts and so on to derive $k$-week ahead prediction. Hence an inaccurate and badly calibrated initial model's forecasts propagate their errors to subsequent predictions as well. 
\begin{wraptable}[8]{r}{.65\linewidth}
\centering
\caption{Evaluation scores for ARI task.}
\scalebox{0.8}{\small
    \begin{tabular}{|l|r|r|r|r|r|r|r|r|r|}
\hline
                               & \multicolumn{3}{c|}{{\ul \textbf{RMSE}}}                  &  \multicolumn{3}{c|}{{\ul \textbf{LS}}} & \multicolumn{3}{c|}{{\ul \textbf{CS}}} \\ \hline
{\ul \textit{Model}}           & k=2                         & k=3             & k=4             &  k=2             & k=3             & k=4             & k=2                & k=3               & k=4               \\ \hline
\textbf{ED}                    & 2.21                      & 3.13          & 3.82                    & 6.03          & 8.84          & 10         & 0.42             & 0.45            & 0.48            \\ \hline
\textbf{MCDP}                  & 3.62                      & 4.03          & 4.39                   & 10         & 10         & 10            & 0.47             & 0.46            & 0.49            \\ \hline
\textbf{BNN}                   & 3.41 & 4.23          & 4.78                   & 10       & 10         & 10        & 0.39             & 0.41            & 0.42            \\ \hline
\textbf{GP}                    & 1.24                      & 1.31          & 1.38                   & 4.62          & 5.17          & 5.51          & 0.37             & 0.36            & 0.37            \\ \hline
\textbf{\model} & \textbf{0.6}              & \textbf{0.85} & \textbf{0.99}  & \textbf{0.64} & \textbf{0.96} & \textbf{1.14} & \textbf{0.063}   & \textbf{0.074}  & \textbf{0.048}  \\ \hline
\end{tabular}
}
\label{tab:ar}
\end{wraptable}
We perform forecasting for $k=2,3,4$ week ahead as described above using the $k=1$ trained model. The pseudocode for Autoregressive inference is given in the Appendix.


\noindent
\textbf{Results:}  See Table~\ref{tab:ar}. Only baselines not trained autoregressively by default (as \model already outperforms them (Q1)) are considered. \model outperforms all  and is  comparable even to the \textit{non AR trained original} \model scores (Table~\ref{tab:kddpred}) whereas we observed a significant deterioration in scores for other baselines, as anticipated.

\vspace{-5pt}
\subsection{Q4: Reacting to abnormal/novel patterns}
\label{sec:adapt}
\noindent  
\textbf{Motivation:} A major challenge in real-time epidemiology~\cite{rodriguez_steering_2020} is the presence of novel patterns e.g. consider the impact of the COVID-19 pandemic on the 2019/20 wILI values (see Figure~\ref{fig:2seasons}). In such cases, a trustworthy real-time forecasting model to anticipate, quantify and adapt is needed to such abnormal  situations. 
We studied our performance for the 2009/10 and 2019/20 seasons, which are well known abnormal seasons (due to the H1N1 strain and the COVID-19 pandemic respectively). While we discuss results for $k=3$ week ahead forecasting of  2019/20 season, the results for 2009/10 season and for $k=1,2,4$ lead to similar conclusions.

\begin{wrapfigure}[14]{r}{.6\linewidth}
\vspace{-10pt}
    \centering
    \begin{subfigure}{.48\linewidth}
      \centering
      \includegraphics[width=\linewidth]{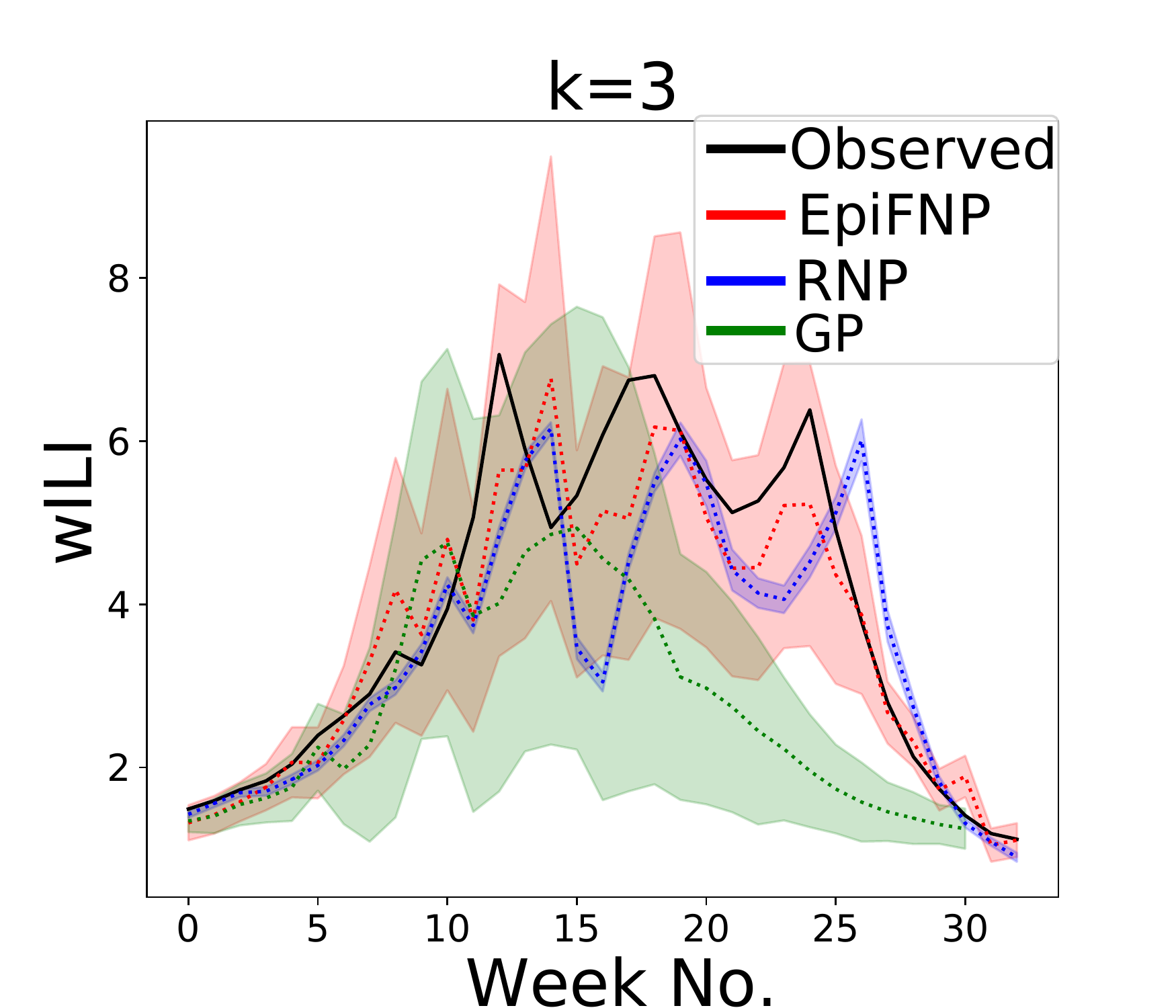}
      \caption{2019/20 predictions}
    \end{subfigure}
    \begin{subfigure}{.48\linewidth}
      \centering
      \includegraphics[width=\linewidth]{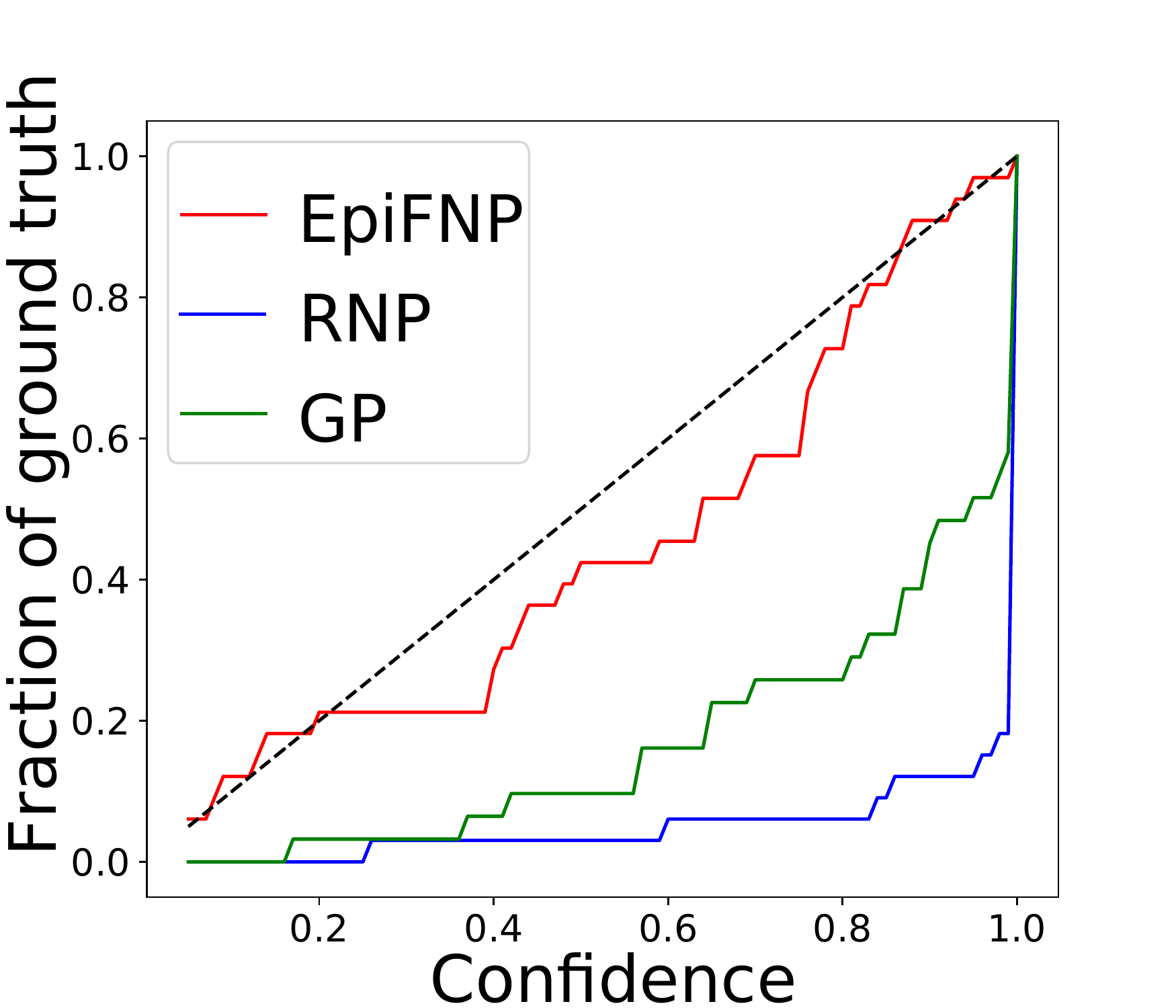}
      \caption{2019/20 CP}
\end{subfigure}
    \caption{\textit{\model outperforms top 2 baselines during abnormal COVID-19 season 2019/20.}}
    \label{fig:2009pred}
\end{wrapfigure}
\noindent 
\textbf{Results:} In short, \textit{\model reacts reliably and adapts to novel scenarios}.
\model outperforms other baselines in all metrics. We observed 18\% and 31\% reduction in RMSE and MAPE respectively compared to best baseline (RNP) and 3.7 times lower LS compared to best baseline (GP). Figure \ref{fig:2009pred}(a) shows the prediction and uncertainty bounds of \model and top 2 baselines.  GP and most other baselines (except RNP) fail to capture the unprecedented third peak around week 24. 
Calibration Plot in Figure \ref{fig:2009pred}(b) shows that \model is better calibrated.

\vspace{-5pt}
\subsection{Q5: Explainable Predictions}
\label{subsec:explainable}
\textbf{Motivation:} Lack of explainability is a major challenge in many ML models, which becomes even more acute in critical domains like public health. Since the Stochastic data correlation graph (SDCG) of \model (recall Section \ref{sec:relation}) explicitly learns to relate each test sequence with relevant historical seasons' sequences, we can leverage this to provide useful explanations for predictions and model uncertainty. Knowing which past seasons are similar is very helpful for epidemiological understanding of the prevalent strain behavior~\cite{adhikari_epideep_2019}. 
We sample SDCGs multiple times and compute average edge probability for every edge between each given historical season and test sequences during real-time forecasting for all weeks. We perform this for $k=3$ weeks ahead forecasting on season 2015/16 but the observations hold for other seasons and $k=1,2,4$ too. 

\begin{wrapfigure}[13]{l}{.55\linewidth}
\vspace{-15pt}
 \centering
\begin{minipage}{.45\linewidth}
      \centering
      \includegraphics[width=\linewidth]{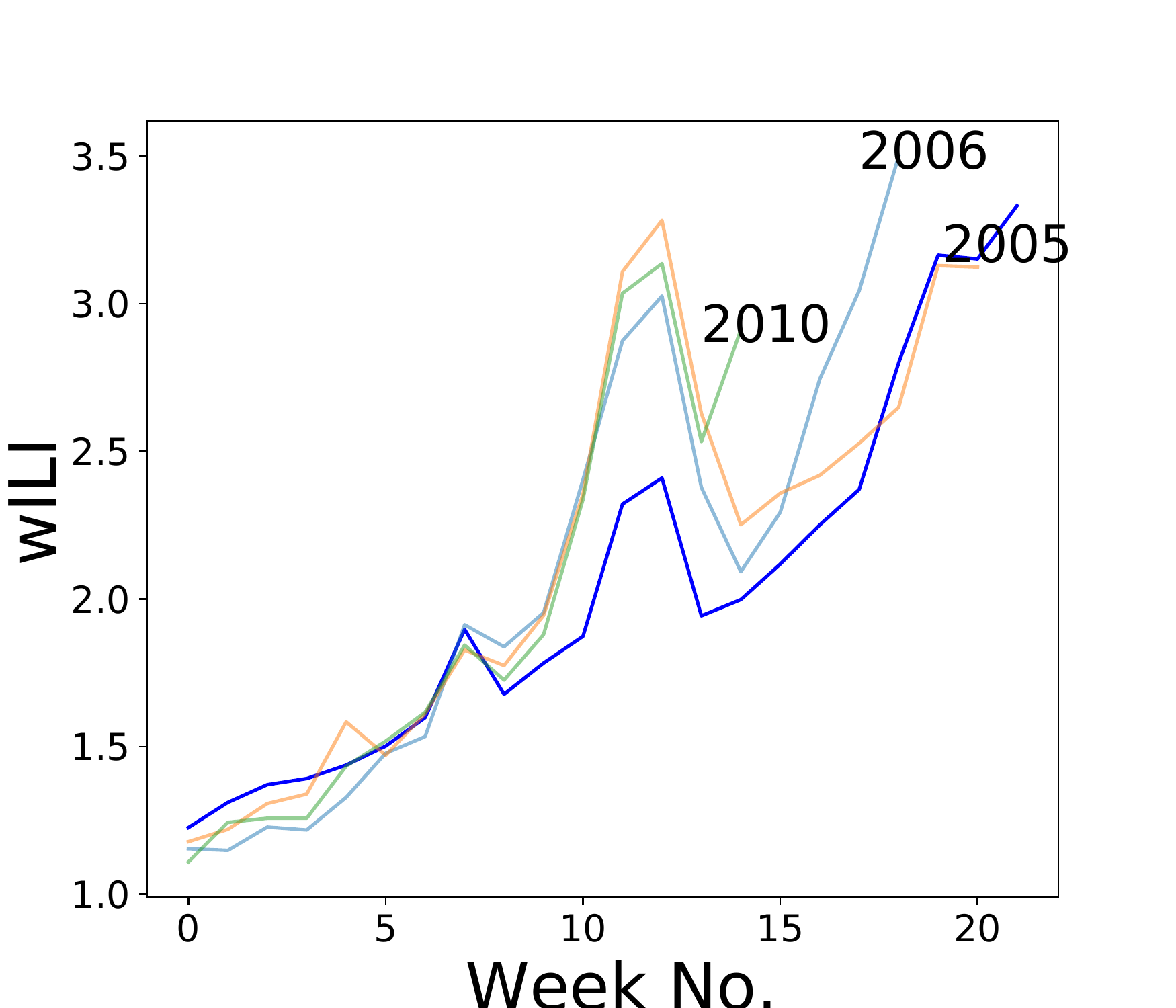}
      \caption{\textit{2015/16 snippet  \& most similar seasons chosen by \model.}}
      \label{fig:2018sim}
    \end{minipage}\hfill
    \begin{minipage}{.52\linewidth}
      \centering
      \includegraphics[width=\linewidth]{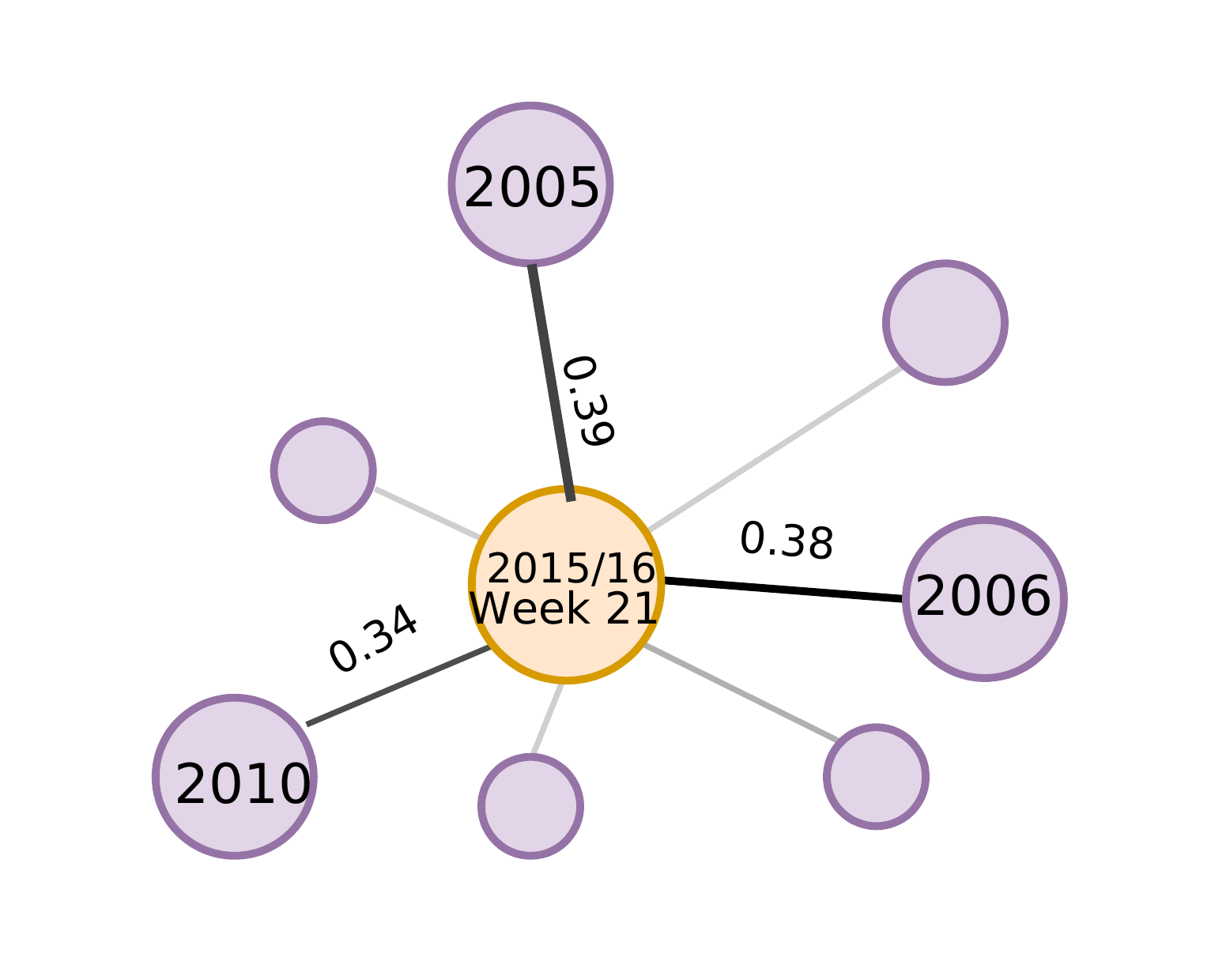}
      \caption{\textit{Average edge probabilities for week 21 of 2015/16 season.}}
      \label{fig:2018sim2}
    \end{minipage}
\end{wrapfigure}
\vspace{-2pt}
\emph{Obs 1: \model automatically chooses most similar historical seasons relevant at time of prediction.}\\
\noindent We leverage the edge probabilities from the SDCG to examine the seasons that are more likely sampled at at each week. We observed that the seasons with higher probabilities showed similar patterns to that of the current test sequence. 
Consider week 21 of season 2015/16 during 3 weeks ahead forecasting. The most likely sampled seasons are  2005, 2006 and 2010 (Figure \ref{fig:2018sim2}). Figure \ref{fig:2018sim} shows these seasons and 2015/16 snippet; clearly they have very similar wILI patterns.




\emph{Obs 2: EpiFNP explains uncertainty bounds of predictions via distribution probabilities in the SDCG.}
 
\begin{wrapfigure}[11]{r}{.55\linewidth}
\vspace{-15pt}
\centering
\begin{minipage}{.48\linewidth}
      \centering
      \includegraphics[width=\linewidth]{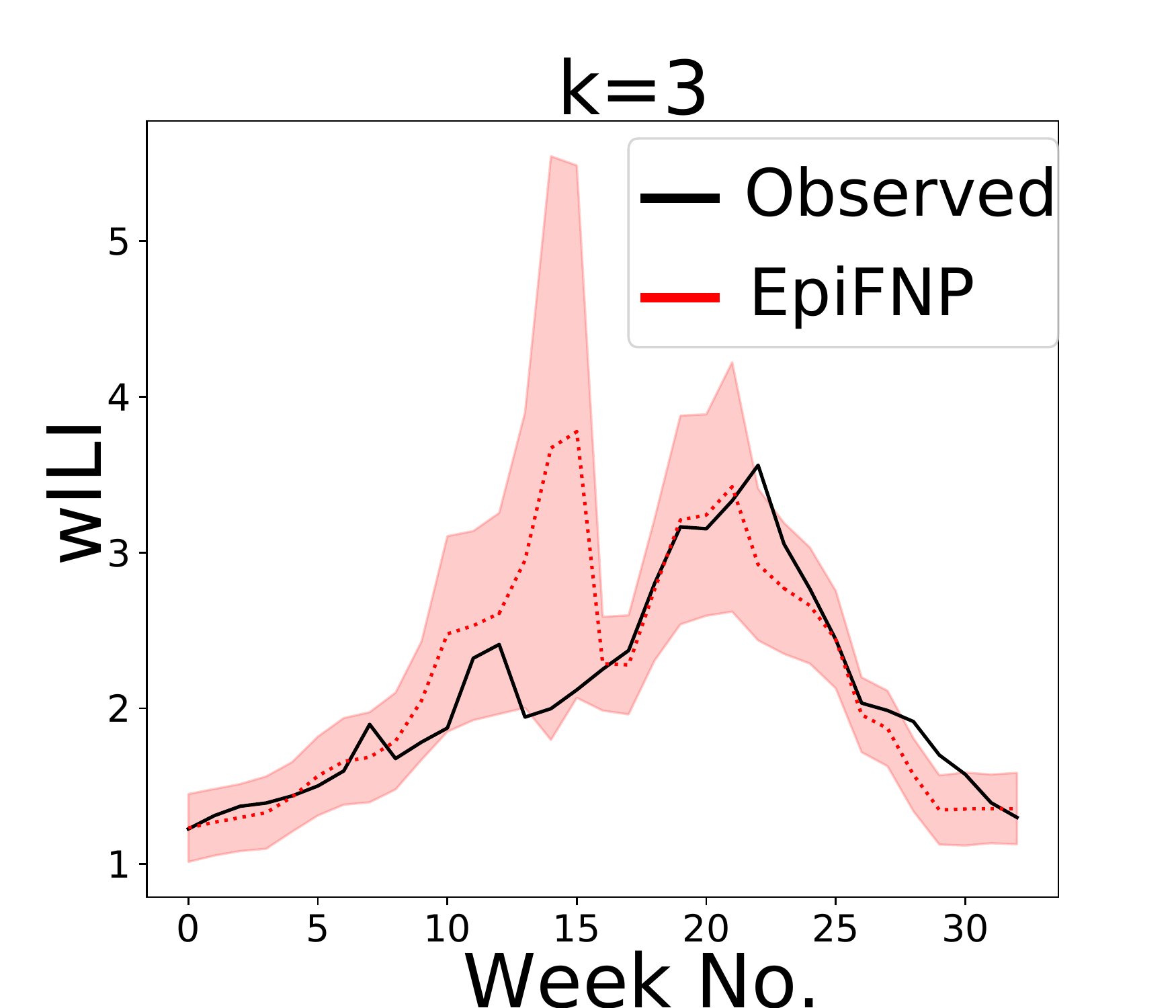}
      \caption{\textit{Higher uncertainty around peaks}}
      \label{fig:2018uncertain1}
    \end{minipage}\hfill
    \begin{minipage}{.51\linewidth}
      \centering
      \includegraphics[width=\linewidth]{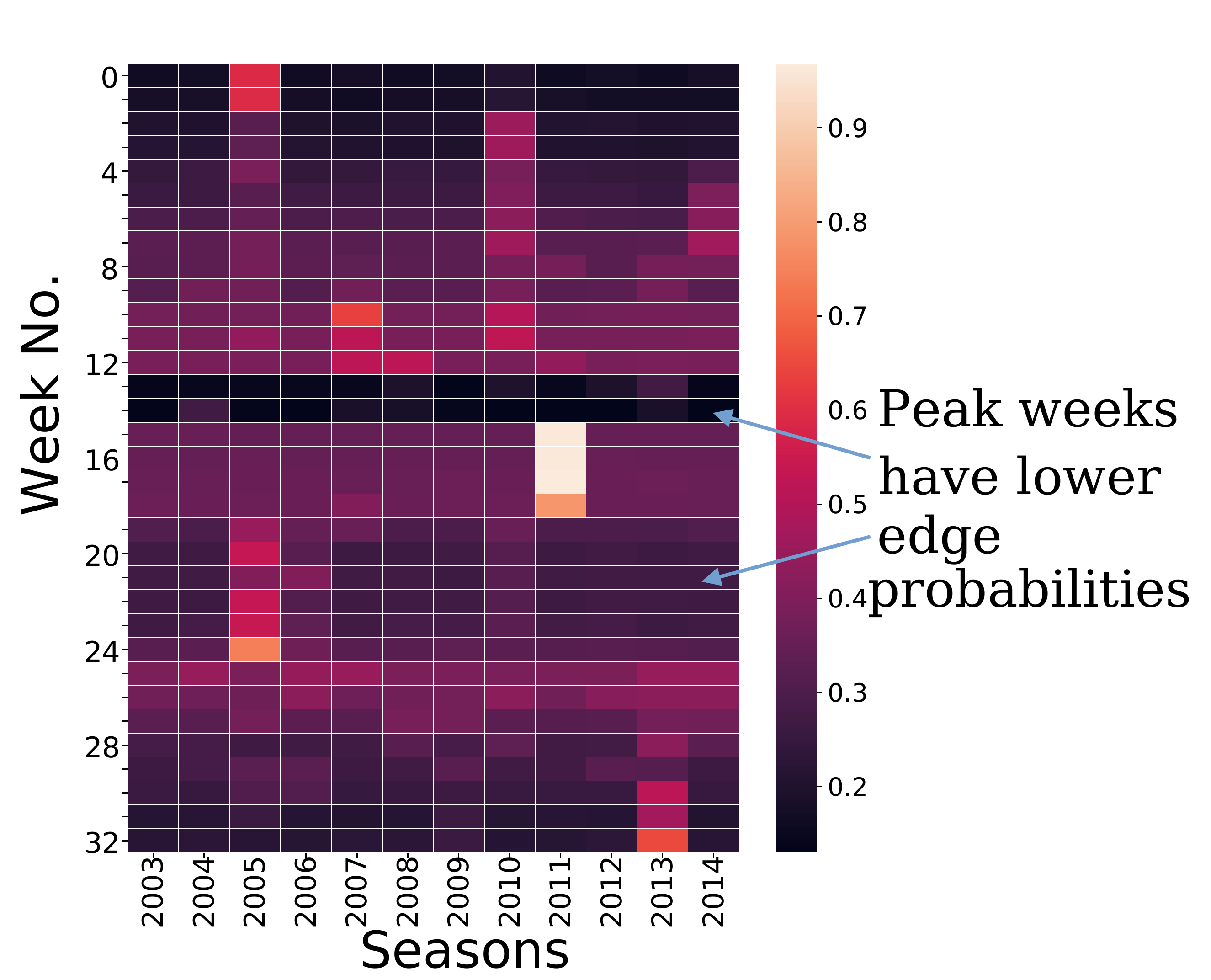}
      \caption{\textit{SDCG Avg. edge probabilities}}
      \label{fig:2018uncertain2}
    \end{minipage}
\end{wrapfigure} 
\noindent 
As seen in Section~\ref{sec:adapt}, \model reacts reliably to abnormal situations and changing trends (e.g. around peaks) by producing larger uncertainty bounds around those events.  
For example, in Figure \ref{fig:2018uncertain1}, uncertainty estimates around peak weeks 12 and 22 are higher than for rest of the weeks.
To examine the source of changing uncertainty bounds of prediction, we look at average edge probabilities generated in SDCG (Figure \ref{fig:2018uncertain2}) and find that around the peak weeks the edge probabilities are lower than in surrounding weeks.
This promotes larger variety of small subsets of the reference set to be sampled during inference that increases the variance of local latent variable $\mathbf{z}_{i}^M$ thereby increasing the variance of the output forecast distribution. 

\vspace{-8pt}
\section{Conclusion}
We introduced \model, a novel deep probabilistic sequence modeling method which generates \emph{well calibrated}, \emph{explainable} and \emph{accurate} predictions. We demonstrated its superior performance in the problem of real-time influenza forecasting by significantly outperforming other non-trivial baselines (more than 2.5x in accuracy and upto 2.4x in calibration). 
Importantly, it was the only one capable of reliably handling unprecedented scenarios e.g. H1N1 and COVID19 seasons. We also showcased its explainability as it automatically retrieves the most relevant historical sequences matching its current week's predictions using the SDCG. All these highlight the usefulness of \model for the complex challenge of trustworthy epidemiological forecasting, which directly impacts public health policies and planning. 
However \model
can be affected by any systematic biases in data collection (for example, some regions might have poorer surveillance and reporting capabilities). There is limited potential for misuse of our algorithms
and/or data sources though the dataset is public/anonymized without any sensitive patient information.

We believe our work opens up many interesting future questions. Our setup can be
easily extended to handle other diseases 
and our core technique can
be adapted for other general sequence modeling problems. Further, we can extend
\model to also use heterogeneous data from multiple sources. We can also explore
incorporating domain knowledge of prior dependencies between different
sources/features (e.g.  geographically close regions are more likely to have
similar disease trends).  


\textbf{Acknowledgments:} We thank the anonymous reviewers for their useful
comments. This work was supported in part by the NSF (Expeditions CCF-1918770,
CAREER IIS-2028586, RAPID IIS-2027862, Medium IIS-1955883, Medium IIS-2106961,
CCF-2115126, Small III-2008334), CDC MInD program, ORNL, ONR MURI
N00014-17-1-2656, faculty research awards from Facebook, Google and Amazon and
funds/computing resources from Georgia Tech.

\printbibliography

\newpage
\appendix
\textbf{\fontsize{15}{12}\selectfont Appendix for When in Doubt: Neural Non-Parametric Uncertainty Quantification for Epidemic Forecasting}

Code for \model  and wILI  dataset is publicly available \footnote{Link to code and dataset: \url{https://github.com/AdityaLab/EpiFNP}}.

\section{Additional Related work}
\textbf{Statistical models for Epidemic Forecasting}
In the recent years, statistical models have been the most successful in several forecasting targets, as noted in multiyear assessments ~\cite{reich_collaborative_2019}. In influenza forecasting, various recent statistical approaches have been proposed. 
On one hand, we have models designed to model the details on the underlying generative distribution of the data. Among these, \cite{brooks_flexible_2015} proposed a semiparametric Empirical Bayes framework that constructs a prior of the current season's epidemic curve from the past seasons and outputs a distribution over epidemic curves. \cite{brooks_nonmechanistic_2018} opts for a non-parametric approach based on kernel density estimation to model the probability distribution of the change between consecutive predictions. Closely related, Gaussian processes have been recently explored for influenza forecasting~\cite{zimmer_influenza_2020}. Other popular methods rely on ensembles of mechanistic and statistical methods ~\cite{ray_infectious_2017}. 

More recently, the deep learning community has take interest in forecasting influenza \cite{adhikari_epideep_2019,wang2019defsi} and COVID-19 \cite{rodriguez_deepcovid_2021}. Indeed, deep learning enables to address novel situations where traditional influenza models fail such as adapting a historical influenza model to pandemic \cite{rodriguez_steering_2020}. Deep learning is also suitable because it provides the capability of ingesting data from multiple sources, which better informs the model of what is happening on the ground. However, for most of this body of work uncertainty quantification is either non existent or has been explored with simple techniques that lack of proper knowledge representation. Our work aims to close this gap in the literature. 

\noindent\textbf{Uncertainty Quantification for Deep Learning} Recent works have shown that deep neural networks are over-confident in their predictions \cite{guo2017calibration, kong2020calibrated}.
Existing approaches for uncertainty quantification  can be categorized into three lines. The first line is based on 
Bayesian Neural Networks (BNNs) \cite{mackay1992practical, blundell2015weight, louizos2017multiplicative}. They are realized by first imposing prior distributions over neural network parameters, then infer parameter posteriors
and further integrate over them to make predictions. However, as exact inference of parameter posteriors is often
intractable, approximation methods have also been proposed, including variational inference \cite{blundell2015weight, louizos2017multiplicative}, Monte
Carlo dropout \cite{gal_dropout_2016} and stochastic gradient Markov chain Monte Carlo (SG-MCMC) \cite{li2016preconditioned, Zhang2020Cyclical}.
Such BNN approximations tend to underestimate the uncertainty \cite{kong2020sde}. Moreover, specifying parameter priors for BNNs is challenging because the parameters of DNNs are huge in size and uninterpretable \cite{kong2020sde, louizos_functional_2019}. 

The second line tries to combine the stochastic processes and DNNs. Neural Process (NP)
\cite{garnelo_neural_2018} defines a distribution over a global latent variable to capture the functional uncertainty, while Functional neural process (FNP) \cite{louizos_functional_2019}  use a dependency graph to encode the data correlation uncertainty. However, they are both for the static data. Recently, recurrent neural process (RNP) \cite{qin_recurrent_2019,kim2019attentive} has been proposed to incorporate RNNs into the NP to capture the ordering sequential information.

The third line is based on model ensembling \cite{lakshminarayanan2017simple} which trains multiple DNNs with different initializations and use their predictions for uncertainty quantification. However, training multiple DNNs require extensive computing resources.

\section{Model Hyperparameters}
We describe all the hyperparameters used for the \model model including the model architecture. In general, we used the hyperparameters as done in \cite{louizos_functional_2019} with changes made to accommodate the sequential modules and global embedding for our use case.
\subsection{Architecture}
\subsubsection{Probabilistic Neural Sequence Encoder}
The GRU for the encoder model has single hidden layer of 50 units and outputs 50 dimensional vectors. The Attention layer was similar to that used in transformers. We used a single attention head and retained the same number of dimensions, 50, when generating the key and value embeddings. to generate $\mathbf{u}_{i}$, we derived the mean and log variance using a stack of 3 linear layers for $g_1$ and $g_2$ with ReLU in between the hidden layers. All hidden layers have 50 units.

Note that for sampling from multivariate gaussian distribution, we always assumed the covariance matrix to be a diagonal matrix and only derived log variance for each dimension.

\subsubsection{Parameterizing Predictive Distribution}
The $h_1$ and $h_2$ functions used to derive $\mathbf{z}_i^M$ were single linear layers with no activation function. The Attention layer used to derive $\mathbf{v}$ was similar to that used in encoder: 1 attention head with 50 dimension units for key and value transforms. $d_1$ and $d_2$ are  two modules of feed forward layers with a ReLU function between them with first layer having 50 units and the second having 2 to output mean and log variance of forecast output.

\subsection{Other Hyperparameters}
Learning rate used was $1e-4$. We also used early stopping to prevent overfitting and randomly sampled 5\% of training points as validation set to determine when we reached the point of overfitting. \model usually 2000-3000 epoch to complete training. We found that our model was very robust to small changes in architecture and learning rate and we mostly optimized for faster rate of convergence during training.

\section{Details on Evaluation metrics}
Let $x_{N+1}^{1...t}$ be a given partial wILI test sequence with observed ground truth $y_{N+1}^{(t)}$ i.e., for a $k$-week-ahead task $y_{N+1}^{(t)}$ is just $x_{N+1}^{(t+k)}$.
For a model/method $M$ let $\hat{p}_{N+1,M}^{(t)}(Y)$ be the output distribution of the forecast with mean $\hat{y}_{N+1,M}^{(1...t)}$. Then we define the evaluation metrics as follows. We evaluate all the methods based on metrics for measuring prediction accuracy (RMSE, MAPE and LS are commonly used in CDC challenges~\cite{adhikari_epideep_2019, reich_collaborative_2019}) as well as targeted ones (CS) measuring the quality of prediction \textit{calibration} of uncertainty. For all metrics, lower is better.
\model is carefully designed to generate both accurate and well calibrated forecasts, unlike past work which focuses typically on accuracy only.

\noindent$\bullet$  $\mathbf{Root~Mean~Sq.~Error~RMSE}(M) = \sqrt{\frac{1}{T}\sum_{t=1}^{T}(y_{N+1}^{(t)} - \hat{y}_{N+1,M}^{(t)})^2}$\\

    \noindent$\bullet$ $ \mathbf{Mean~Abs.~Per.~Error~MAPE}(M) = \frac{1}{T}\sum_{t=1}^T \frac{|y_{N+1}^{(t)} - \hat{y}_{N+1,M}^{(t)}|}{|y_{N+1}^{(t)}|}$\\

    \noindent$\bullet$ \textbf{Log Score} (LS): This score used by the CDC caters to the stochastic aspect of forecast prediction~\cite{reich_collaborative_2019}.

    \begin{equation}
         LS(M) = \sum_{i=1}^T \frac{1}{T}\int_{y_{i}^{( t)} - 0.5}^{y_{i}^{(t)} + 0.5} - \log (\hat{p}_{N+1,M}^{(t)}(y))dy
    \end{equation}

The integral is approximated by samples from $\hat{p}_{N+1,M}^{(t)}(Y)$ and calculating the fraction of samples that fall in the correct interval.

\noindent$\bullet$ \textbf{Calibration Score} (CS):
In order to evaluate the calibration of output distribution we introduce a new metric called Calibration Score, which is inspired by reliability diagrams \cite{niculescu-mizil_predicting_2005} used for binary events. The idea behind the calibration score is that \emph{a well calibrated model provides meaningful confidence intervals}.
For a model $M$ we define a function $k_M: [0,1] \rightarrow [0,1]$ as follows. For each value of confidence $c \in[0,1]$, let $k_M(c)$ denote the fraction of observed ground truth that lies inside the $c$ confidence interval of predicted output distributions of $M$.
 For a perfectly calibrated model $M^*$ we would expect $k_{M^*}(c)=c$. CS measures the deviation of $k_{M}$ from $k_{M^*}$. Formally, we define CS as:
\begin{equation}
    CS(M)=\int_{0}^1 |k_M(c)-c| dc \approx 0.01 \sum_{c\in\{0, 0.01, \dots, 1\}} |k_M(c)-c|
\end{equation}
(since integrating over all values of $c$ is intractable in general).

We also define the \textbf{Calibration Plot} (CP) as the profile of $k_M(c)$ vs $c$ for all $c\in[0,1]$.

\section{Detailed forecast results}

\subsection{Regional forecasts}
We also evaluate our model and baselines  on wILI dataset specific to different regions in USA. The wILI data for 10 HHS regions are available separately and each of them have different characteristics in their wILI trends which are affected by local climate, population density and other factors. Therefore we train our models on each of the HHS regions seperately and average the scores to produce the results in table \ref{tab:kddregion}. \model outperforms baselines in most baselines. We observed that for 8 of the 10 regions \model outperforms all models in all the evaluation metrics across 2 to 4 week ahead forecast tasks. Even for the remaining 2 regions \model shows  superior scores in majority of the metrics.

\begin{table}[h]
\centering
\caption{Average Evaluation scores of \model and baselines across all HHS regions. The scores are averaged over seasons 2014-15 to 2019-20 for all 10 HHS regions.}
\label{tab:kddregion}
\scalebox{0.8}{
\begin{tabular}{|l|r|r|r|r|r|r|r|r|r|r|r|r|}
\hline
                               & \multicolumn{3}{c|}{{\ul \textbf{RMSE}}}     & \multicolumn{3}{c|}{{\ul \textbf{MAPE}}}      & \multicolumn{3}{c|}{{\ul \textbf{LS}}}        & \multicolumn{3}{c|}{{\ul \textbf{CS}}}        \\ \hline
{\ul Model}                    & 2             & 3            & 4             & 2             & 3             & 4             & 2             & 3             & 4             & 2             & 3             & 4             \\ \hline
\textbf{ED}                    & 0.86          & 1.2          & 1.81          & 0.23          & 0.25          & 0.36          & 2.89          & 2.69          & 3.32          & 0.17          & 0.32          & 0.33          \\ \hline
\textbf{GRU}                   & 1.95          & 2.05         & 2.76          & 0.39          & 0.41          & 0.43          & 4.41          & 4.52          & 4.86          & 0.37          & 0.38          & 0.41          \\ \hline
\textbf{MCDP}                  & 3.01          & 3.36         & 3.41          & 0.58          & 0.548         & 0.68          & 10         & 10         & 10        & 0.38          & 0.39          & 0.47          \\ \hline
\textbf{GP}                    & 0.64          & 0.83         & 0.95          & 0.19          & 0.22          & \textbf{0.25} & \textbf{0.92} & 1.44          & 1.63          & 0.13          & 0.16          & 0.15          \\ \hline
\textbf{BNN}                   & 2.25         & 2.87         & 3.02          & 0.26          & 0.29          & 0.35          & 8.31          & 9.89            & 10            & 0.38         & 0.42          & 0.46          \\ \hline
\textbf{SARIMA}                & 1.81          & 2.33         & 2.8           & 0.36          & 0.47          & 0.58          & 3.3           & 3.87          & 4.37          & 0.39          & 0.37          & 0.37          \\ \hline
\textbf{RNP}                   & 0.87          & 0.88         & 1.17          & 0.19          & 0.23          & 0.29          & 9.27          & 9.58          & 9.78          & 0.46          & 0.46          & 0.47          \\ \hline
\textbf{EB}                    & 1.51          & 1.53         & 1.56          & 0.67          & 0.67          & 0.68          & 7.15          & 7.23          & 7.29          & 0.13          & 0.13          & 0.13          \\ \hline
\textbf{DD}                    & 0.84          & 1.05         & 1.22          & 0.44          & 0.49          & 0.55          & 3.51          & 3.77          & 3.91          & \textbf{0.11} & \textbf{0.11} & \textbf{0.12} \\ \hline
\textbf{\model} & \textbf{0.55} & \textbf{0.7} & \textbf{0.89} & \textbf{0.17} & \textbf{0.19} & 0.26          & 1.41          & \textbf{1.54} & \textbf{1.81} & 0.15          & \textbf{0.11} & 0.13          \\ \hline
\end{tabular}
}

\end{table}

\subsection{Post-hoc calibration methods}
We also evaluated effect of post-hoc methods \cite{kuleshov2018accurate,song2019distribution} on calibration of prediction distributions of top baselines and \model. The results are summarized in Table \ref{tab:posthoc}.
We observe that \model doesn't benefit much from post-hoc calibration methods due to its already well-calibrated forecasts. However, they improve the calibration scores of other baselines (sometimes at the cost of prediction accuracy). However, \model is still clearly the best performing model.

\begin{table}[h]
\caption{Effect of post-hoc calibration on point estimate and calibration scores. Iso and DC are the post-hoc methods introduced in \cite{kuleshov2018accurate} and \cite{song2019distribution} respectively.}
\label{tab:posthoc}
\scalebox{0.8}{
\begin{tabular}{|l|l|l|l|l|l|l|l|l|l|l|l|l|l|}
\hline
                                  &          & \multicolumn{3}{c|}{{\ul \textbf{RMSE}}}      & \multicolumn{3}{c|}{{\ul \textbf{MAPE}}}         & \multicolumn{3}{c|}{{\ul \textbf{LS}}}        & \multicolumn{3}{c|}{{\ul \textbf{CS}}}           \\ \hline
{\ul Model}                       & Post-Hoc & k=2           & k=3           & k=4           & k=2            & k=3            & k=4            & k=2           & k=3           & k=4           & k=2            & k=3            & k=4            \\ \hline
\multirow{3}{*}{\textbf{\model}}  & None     & \textbf{0.48} & \textbf{0.79} & \textbf{0.78} & \textbf{0.089} & \textbf{0.128} & \textbf{0.123} & \textbf{0.56} & \textbf{0.84} & \textbf{0.89} & \textbf{0.068} & \textbf{0.081} & \textbf{0.035} \\ \cline{2-14} 
                                  & Iso      & \textbf{0.49} & \textbf{0.81} & \textbf{0.79} & \textbf{0.09}  & \textbf{0.124} & \textbf{0.119} & \textbf{0.56} & \textbf{0.86} & \textbf{0.9}  & \textbf{0.08}  & \textbf{0.09}  & \textbf{0.07}  \\ \cline{2-14} 
                                  & DC       & \textbf{0.44} & \textbf{0.74} & \textbf{0.77} & \textbf{0.088} & \textbf{0.114} & \textbf{0.117} & \textbf{0.55} & \textbf{0.75} & \textbf{0.86} & \textbf{0.07}  & \textbf{0.08}  & \textbf{0.035} \\ \hline
\multirow{3}{*}{\textbf{RNP}}     & None     & 0.61          & 0.98          & 1.18          & 0.13           & 0.22           & 0.29           & 3.34          & 3.61          & 3.89          & 0.43           & 0.38           & 0.34           \\ \cline{2-14} 
                                  & Iso      & 1.77          & 2.26          & 2.18          & 0.18           & 0.27           & 0.28           & 2.55          & 2.62          & 3.12          & 0.18           & 0.23           & 0.24           \\ \cline{2-14} 
                                  & DC       & 1.73          & 2.17          & 2.25          & 0.18           & 0.27           & 0.31           & 1.53          & 1.84          & 2.05          & 0.13           & 0.12           & 0.15           \\ \hline
\multirow{3}{*}{\textbf{GP}}      & None     & 1.28          & 1.36          & 1.45          & 0.21           & 0.22           & 0.26           & 2.02          & 2.12          & 2.27          & 0.24           & 0.25           & 0.28           \\ \cline{2-14} 
                                  & Iso      & 2.24          & 2.51          & 2.72          & 0.34           & 0.34           & 0.38           & 1.97          & 2.13          & 2.16          & 0.094          & 0.12           & 0.11           \\ \cline{2-14} 
                                  & DC       & 2.15          & 2.68          & 2.72          & 0.32           & 0.37           & 0.39           & 1.94          & 2.07          & 2.04          & 0.09           & 0.11           & 0.1            \\ \hline
\multirow{3}{*}{\textbf{EpiDeep}} & None     & 0.73          & 1.13          & 1.81          & 0.14           & 0.23           & 0.33           & 4.26          & 6.37          & 8.75          & 0.24           & 0.15           & 0.42           \\ \cline{2-14} 
                                  & Iso      & 1.02          & 1.25          & 1.94          & 0.16           & 0.24           & 0.34           & 2.46          & 4.58          & 4.64          & 0.21           & 0.11           & 0.19           \\ \cline{2-14} 
                                  & DC       & 1.15          & 1.28          & 1.74          & 0.17           & 0.26           & 0.32           & 2.11          & 3.97          & 3.65          & 0.18           & 0.14           & 0.21           \\ \hline
\multirow{3}{*}{\textbf{MCDP}}    & None     & 2.24          & 2.41          & 2.61          & 0.46           & 0.51           & 0.6            & 9.62          & 10            & 10            & 0.24           & 0.32           & 0.34           \\ \cline{2-14} 
                                  & Iso      & 2.36          & 2.58          & 2.53          & 0.45           & 0.47           & 0.59           & 6.72          & 9.64          & 10            & 0.14           & 0.26           & 0.31           \\ \cline{2-14} 
                                  & DC       & 2.31          & 2.44          & 2.52          & 0.44           & 0.48           & 0.57           & 6.31          & 8.24          & 10            & 0.15           & 0.22           & 0.25           \\ \hline
\end{tabular}
}
\end{table}

\section{Autoregressive inference}
We formally describe how to perform autoregressive inference as discussed in Section \ref{sec:ar} in Algorithm \ref{alg:AR}.
\begin{algorithm}[thp]
{\small 
\SetAlgoLined
\SetKwInOut{Input}{Input}
\SetKwInOut{Output}{Output}
\Input{Model $M$ trained for 1 week ahead forecasting, test sequence $x_{i}^{(1...t)}$, $k$: No. of weeks ahead to forecast}
\Output{ Distribution $\hat{P}_M(X_{i}^{(t+k)}|x_{i}^{(1...t)})$ for forecasting $x_{i}^{(t+k)}$}

\tcc{$Z_i$ is the set of candidate sequences for $t+i+1$ forecasting.  Each sequence has first $t$ values as $x_{t}^{1...t}$ and next $i$ values are sampled by ARI }
$Z_0 = \{x_{i}^{(1...t)}\}$\;
\For{$i$ in 1 to $k$}{
    \For{$j$ in 1 to $N$}{
        Sample sequence $\bar{x}$ from $Z_{i-1}$\;
        Feed $\bar{x}$ to $M$ and sample output $y$\;
        Append $y$ to $\bar{x}$ to form a new sequence $\bar{x}\oplus \{y\}$\;
        Add $\bar{x}\oplus \{y\}$ to $Z_i$\;
        \tcp{$\bar{x}\oplus \{y\}$ is a candidate sequence for $t+i+1$ forecast.}
    }
}
$\mathbf{preds}=\{x: x \text{ is last element of } \bar{x}\in Z_k\}$\;
Approximate $\hat{P}_M(X_{i}^{(t+k)}|x_{i}^{(1...t)})$ from $\mathbf{preds}$

\caption{Autoregressive inference (ARI)}\label{alg:AR}}
\end{algorithm}
\subsection{Results}
\begin{figure}[h]
    \centering
    \includegraphics[width=.4\linewidth]{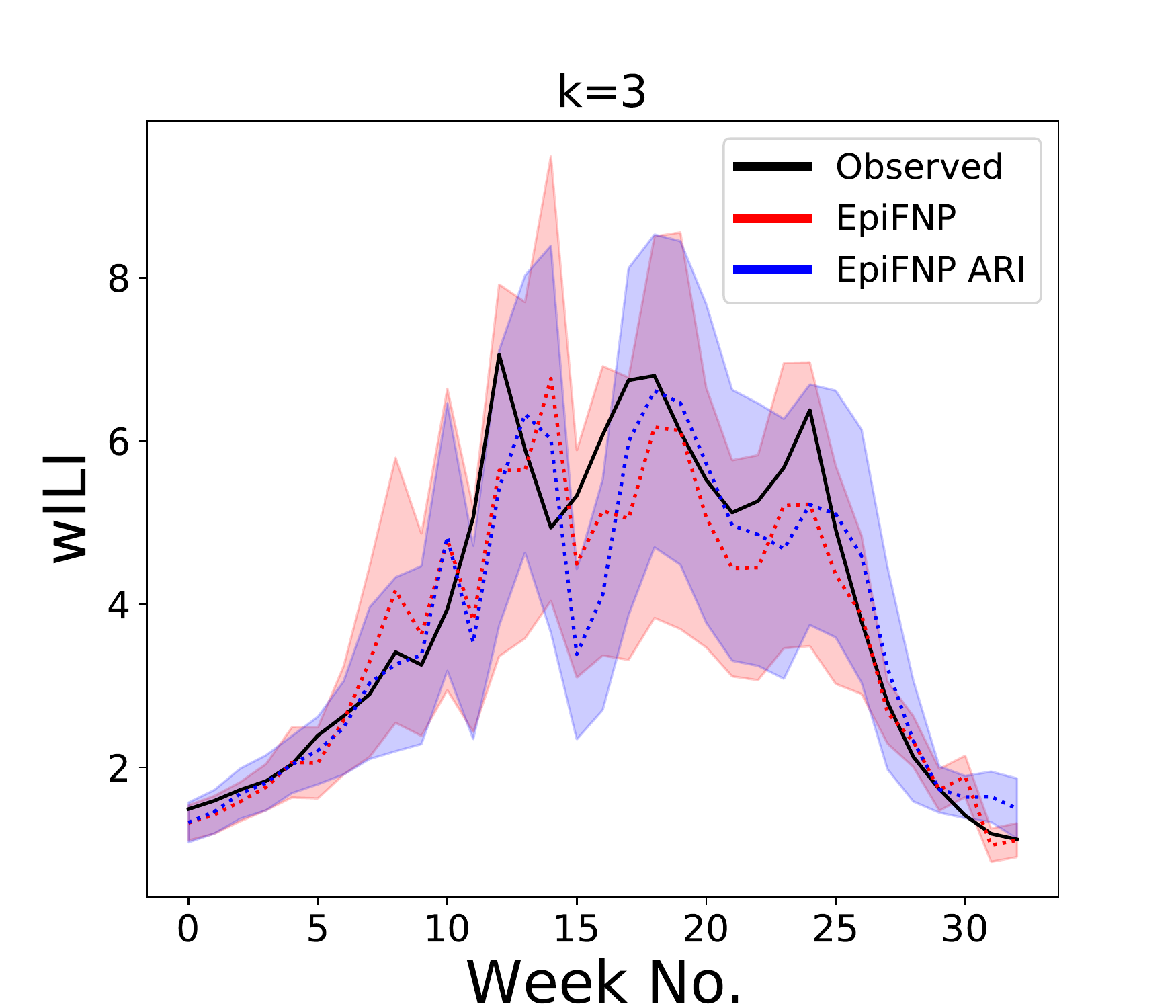}
    \caption{Uncertainty bounds of ARI \model and normally trained \model are similar.}
    \label{fig:app_ari}
\end{figure}

We provided RMSE, LS and CS of AR task in main paper Table 2.
See Table \ref{tab:app_ar} for results for AR task that includes MAPE scores. As described in Section \ref{sec:ar}, \model outperforms baselines in AR tasks and its performance in comparable to \model scores trained separately for different values of $k$ (Figure \ref{fig:app_ari}).
\begin{table}[h]
    \centering
     \caption{Evaluation scores for ARI task (Section \ref{sec:ar})}
    \label{tab:app_ar}
    \scalebox{0.8}{
     \begin{tabular}{|l|r|r|r|r|r|r|r|r|r|r|r|r|}
\hline
                               & \multicolumn{3}{c|}{{\ul \textbf{RMSE}}}                  & \multicolumn{3}{c|}{{\ul \textbf{MAPE}}}      & \multicolumn{3}{c|}{{\ul \textbf{LS}}} & \multicolumn{3}{c|}{{\ul \textbf{CS}}} \\ \hline
{\ul \textit{Model}}           & $k=2$                         & $k=3$             & $k=4$             & $k=2$            & $k=3$             & $k=4$              & $k=2$             & $k=3$             & $k=4$             & $k=2$                & $k=3$               & $k=4$               \\ \hline
\textbf{ED}                    & 2.21                      & 3.13          & 3.82          & 0.4          & 0.43          & 0.55           & 6.03          & 8.84          & 10         & 0.42             & 0.45            & 0.48            \\ \hline
\textbf{MCDP}                  & 3.62                      & 4.03          & 4.39          & 0.58         & 0.61          & 0.67           & 10         & 10         & 10            & 0.47             & 0.46            & 0.49            \\ \hline
\textbf{BNN}                   & 3.41 & 4.23          & 4.78          & 0.51         & 0.55          & 0.62           & 10       & 10         & 10        & 0.39             & 0.41            & 0.42            \\ \hline
\textbf{GP}                    & 1.24                      & 1.31          & 1.38          & 0.21         & 0.21          & 0.24           & 4.62          & 5.17          & 5.51          & 0.37             & 0.36            & 0.37            \\ \hline
\textbf{\model} & \textbf{0.6}              & \textbf{0.85} & \textbf{0.99} & \textbf{0.1} & \textbf{0.14} & \textbf{0.166} & \textbf{0.64} & \textbf{0.96} & \textbf{1.14} & \textbf{0.063}   & \textbf{0.074}  & \textbf{0.048}  \\ \hline
\end{tabular}
}
   
\end{table}

\section{Ablation study}
We examine the effectiveness of three components of \model in learning accurate predictions and good calibration of uncertainty: (1) Global Latent Variable $\mathbf{v}$ , (2) Local latent variable $\mathbf{z}_{i}^M$ (3) Modelling sequence encodings $\mathbf{u_i}$ as a random variable instead of directly using deterministic encodings $\bar{\mathbf{h}}_{i}$.
Detailed results of this study are in Table \ref{tab:kddablation}.
All three components are essential for best performance of the model. Removing $\mathbf{z}_{i}^M$  shows very large decrease in log scores and calibration scores.
This aligns with the hypothesis about role of data correlation graph in determining uncertainty bounds (see Section \ref{subsec:explainable}).

We present the results of ablation experiments in Table \ref{tab:kddablation}.
We see that all three components are essential for best performance of the model. Removing $\mathbf{z}^M_i$ shows large decrease in log scores and calibration scores as the model becomes less capable of modelling uncertainty. Of all the ablation models, making latent embeddings deterministic seems to have least effect on performance though the reduction is still very detrimental to overall performance.

\begin{table}[h]
\centering
\caption{Ablation study to measure the effects of 1) Local latent variable $\mathbf{z}_i^M$ 2)  Global latent variable $\mathbf{v}$ and 3) Stochastic SeqEncoder: Modelling $\mathbf{u_i}$ as stochastic latent variables rather than deterministic encodings.}
\label{tab:kddablation}
\scalebox{0.9}{
\begin{tabular}{|l|r|r|r|r|r|r|}
\hline
Ablation study                     & \multicolumn{3}{l|}{RMSE}                     & \multicolumn{3}{l|}{MAPE}                         \\ \hline
Model/Weeks ahead                  & 2             & 3             & 4             & 2              & 3              & 4               \\ \hline
\model                             & \textbf{0.48} & \textbf{0.79} & \textbf{0.78} & \textbf{0.089} & \textbf{0.128} & \textbf{0.123}  \\ \hline
-(Local latent variable)                              & 0.99          & 1.45          & 1.51          & 0.17           & 0.25           & 0.29                  \\ \hline
-(Global latent variable)                              & 1.76          & 2.05          & 2.45          & 0.33           & 0.41           & 0.42           \\ \hline
- (Stochastic Encoder)                             & 0.87          & 1.09          & 1.19          & 0.15           & 0.21           & 0.22              \\ \hline
-(Stochastic Encoder, Local latent variable)                         & 1.18          & 1.39          & 1.83          & 0.17           & 0.18            & 0.21           \\ \hline
- (Stochastic Encoder, Global latent variable) & 0.67          & 0.73          & 0.9           & 0.19           & 0.2            & 0.26            \\ \hline
Ablation study                      & \multicolumn{3}{l|}{LS}          & \multicolumn{3}{l|}{CS}            \\ \hline
Model/Weeks ahead                   & 2             & 3             & 4            & 2              & 3              & 4              \\ \hline
\model                             & \textbf{0.51} & \textbf{0.78} & \textbf{1.2} & \textbf{0.069} & \textbf{0.081} & \textbf{0.035} \\ \hline
-(Local latent variable)                                   & 3.51          & 6.67          & 8.09         & 0.21           & 0.27           & 0.29           \\ \hline
-(Global latent variable)                                & 2.06          & 2.41          & 3.37         & 0.085          & 0.12           & 0.19           \\ \hline
- (Stochastic Encoder)                              & 3.13          & 3.53          & 4.88         & 0.14           & 0.19           & 0.24           \\ \hline
-(Stochastic Encoder, Local latent variable)                                & 6.11          & 8.91          & 9.68         & 0.44           & 0.48           & 0.47           \\ \hline
- (Stochastic Encoder, Global latent variable)    & 2.21          & 3.58          & 3.72         & 0.41           & 0.45           & 0.42           \\ \hline
\end{tabular}
}
\end{table}

\section{\model adapts to H1N1 Flu season}
\begin{figure}[h]
    \centering
    \begin{subfigure}{.48\linewidth}
      \centering
      \includegraphics[width=\linewidth]{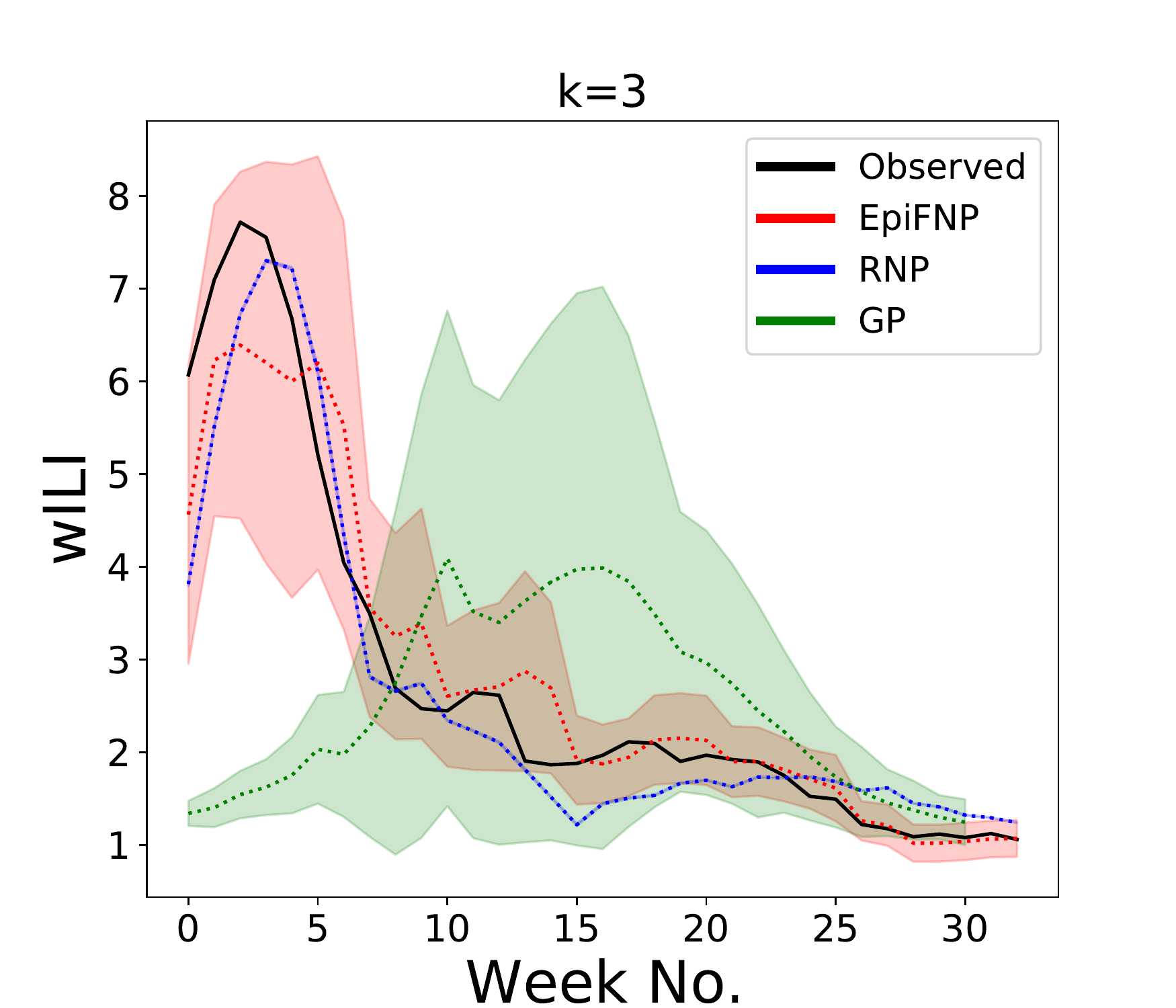}
      \caption{2009/10 predictions}
    \end{subfigure}\hfill
    \begin{subfigure}{.48\linewidth}
      \centering
      \includegraphics[width=\linewidth]{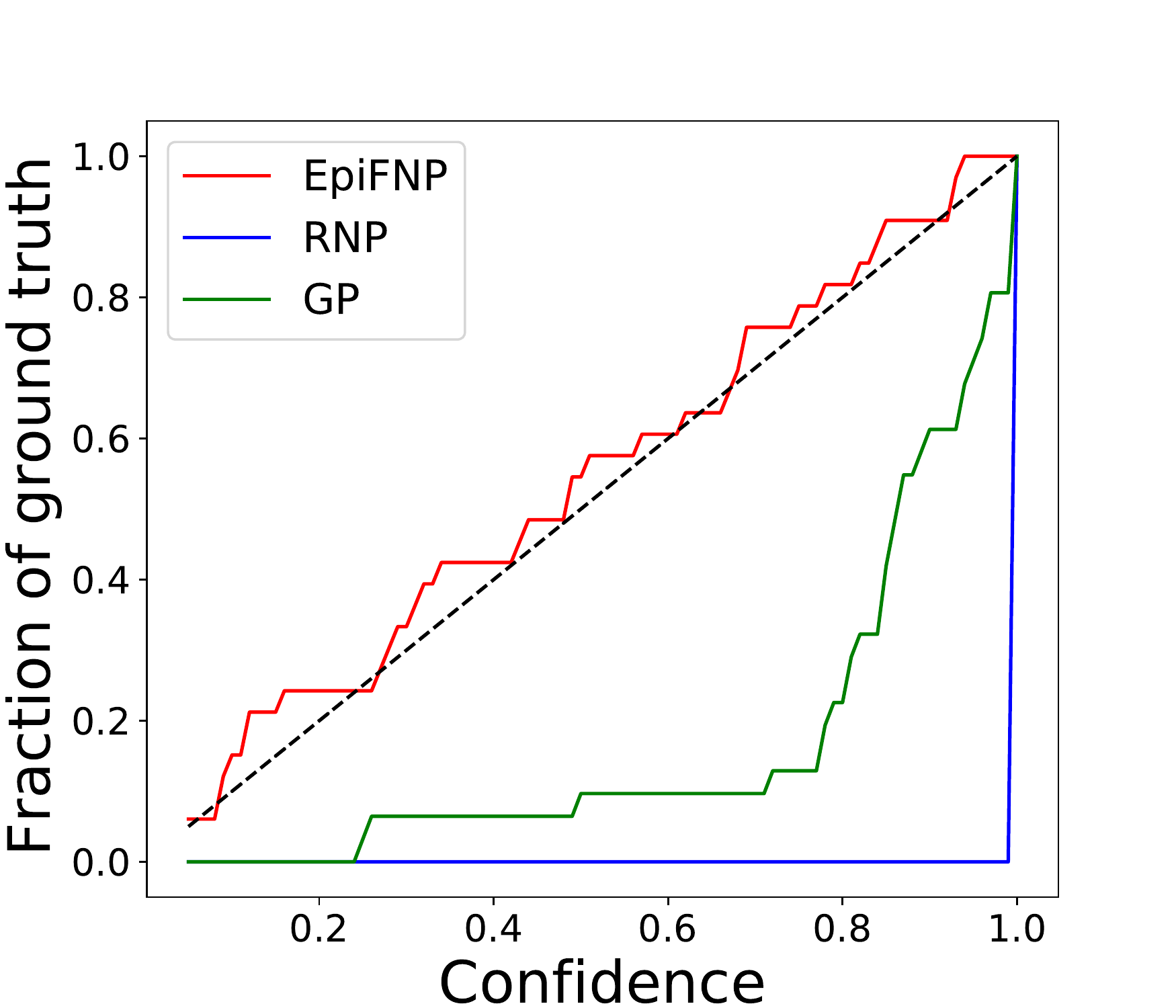}
      \caption{2009/10 Calibration plot}
    \end{subfigure}\hfill
    \caption{\model outperforms baselines on real-time forecasting during abnormal H1N1 season (2009/10). Forecasts for $k=3$ weeks ahead forecast by \model and next two best baselines: RNP and GP.}
    \label{fig:app2009pred}
\end{figure}

 \model outperforms all baselines and has 30\% and 10\% better RMSE and MAPE scores compared to second best baseline (RNP). LS of \model is 0.48, about \textit{9.8 times} lesser than second best model. 
Figure \ref{fig:app2009pred}(a) shows the prediction and 95\% confidence bounds of \model and two best performing baselines. \model captures the unprecedented early peak observed around week 4. There is also a high uncertainty bounds around the peak. In contrast RNP has very small uncertainty bounds. GP and most other baselines (except GRU, RNP and MCDP) do not even capture the peak.
 Calibration plot in Figure \ref{fig:app2009pred}(b) shows the deviation of \model from ideal diagonal to be much smaller compared to other baselines. This results in about \textit{4.6 times} smaller CS compared to the best baseline.

\end{document}